\documentclass[journal]{IEEEtran}
\usepackage{hyperref}
\hypersetup{
    colorlinks=true,
    citecolor=blue
}
\usepackage[T1]{fontenc}
\usepackage[numbers,compress]{natbib}
 
\usepackage{array}
\usepackage{graphicx} 
\usepackage{subfigure}
\usepackage{algpseudocode}  
\usepackage{amsmath}  
\usepackage{bm}
\usepackage{tabularx}
\usepackage{CJKutf8}
\usepackage[cmintegrals]{newtxmath}
\usepackage{url}
\usepackage[usenames,dvipsnames]{color}
\definecolor{Gray}{gray}{0.95}
\definecolor{mygreen}{rgb}{0.1,0.255,0.1}
\usepackage{colortbl}
\usepackage{multirow}
\usepackage[utf8]{inputenc}
\usepackage{cleveref}

\crefname{section}{§}{§§}
\Crefname{section}{§}{§§}
\usepackage{enumitem}
\usepackage{pifont}

\usepackage{ulem}
\usepackage{amsfonts}
\usepackage{algorithm}
\usepackage{algorithmicx}
\usepackage{xspace}

\usepackage{ulem} 

\usepackage{booktabs} 
\usepackage{hhline}   

\usepackage[table]{xcolor} 

\usepackage{tcolorbox}

\usepackage{tikz,xcolor}
\definecolor{lime}{HTML}{A6CE39}
\DeclareRobustCommand{\orcidicon}{
\begin{tikzpicture}
\draw[lime, fill=lime] (0,0)
circle[radius=0.16]
node[white]{{\fontfamily{qag}\selectfont \tiny \.{I}D}}; 

\end{tikzpicture}

\hspace{-2mm}
}
\foreach \x in {A, ..., Z}{%
\expandafter\xdef\csname orcid\x\endcsname{\noexpand\href{https://orcid.org/\csname orcidauthor\x\endcsname}{\noexpand\orcidicon}}
}

\newcommand{\NANONET}{%
  \textsc{\bfseries\color{blue!70!black}NanoNet}%
}

\newcommand\SystemNameA{\textsc{DisCo}\xspace}
\newcommand\SystemNameB{\textsc{PsNet}\xspace}
\newcommand\SystemNameO{\textsc{NanoNet}\xspace}

\definecolor{Gray}{gray}{0.95}

\ifCLASSINFOpdf
\else 
\fi

\begin{document}
\title{\NANONET{}: Parameter-Efficient Learning with Label-Scarce Supervision for Lightweight \\ Text Mining Model}

\author{\mbox{Qianren~Mao\orcidA{}, 
Yashuo~Luo,
Ziqi~Qin,
Junnan~Liu,
Weifeng~Jiang,}\\
\mbox{
Zhijun~Chen\orcidC{}, 
Zhuoran~Li\orcidH{},
Likang~Xiao\orcidJ{},
Chuou~Xu,
Qili Zhang,
Hanwen Hao,
Jingzheng Li\orcidK{},} \\ 
\mbox{Chunghua Lin, 
Jianxin~Li\orcidE{},~\IEEEmembership{Member, IEEE},
Philip~S. Yu\orcidG{},~\IEEEmembership{Life Fellow, IEEE}}
\IEEEcompsocitemizethanks{\IEEEcompsocthanksitem

Recommended for acceptance by xxxx (\textit{Corresponding authors}: Qianren Mao.)
Qianren. Mao (E-mail: maoqr@zgclab.edu.cn), is with the Zhongguancun Laboratory, Beijing 100095, China. 
Qili Zhang (E-mail: 20373496@buaa.edu.cn), Hanwen Hao (E-mail: 20373190@buaa.edu.cn),  Zhijun Chen (E-mail: zhijunchen@buaa.edu.cn), Jianxin Li (E-mail: lijx@buaa.edu.cn) are with the School of Computer Science and Engineering, Beihang University, Beijing 100191, China. 
Weifeng Jiang (E-mail:weifeng001@e.ntu.edu.sg) is with Nanyang Technological University (NTU), 639798, Singapore.   
Chenghua Lin (E-mail: chenghua.lin@manchester.ac.uk) is with the Department of Computer Science, The University of Manchester, UK.
Jinhu Lü (e-mail: ihlu@iss.ac.cn) is with the School of Automation Science and Electrical Engineer-ing, State Key Laboratory of Software Development Environment, Beihang University, Beijing 100191, China.
Jianxin Li, Jinhu Lü are also with the Zhongguancun Laboratory.
Philip~S. Yu (E-mail: psyu@uic.edu) is with the Department of Computer Science, University of Illinois at Chicago, Chicago 60607 USA.

Portions of this work were previously presented in:
Q. Mao, W. Jiang, J. Liu et al., Lightweight Contenders: Navigating Semi-Supervised Text Mining through Peer Collaboration and Self Transcendence. Findings of NAACL 2025, and
W. Jiang, Q. Mao et al, C Lin., DisCo: Distilled Student Models Co-training for Semi-supervised Text Mining. EMNLP 2023.
These papers serve as baselines herein. The present manuscript significantly extends them by (i) \SystemNameO differs from \SystemNameA and \SystemNameB by employs advanced parameter-efficient  training methods that ensure the training process requires only a small amount of labeled data and a limited training  parameters, and that the final inference model is also small. (ii) \SystemNameA and \SystemNameB also use scarce labels and output compact inference models, but they keep the full parameter set trainable, so their trainable count far exceeds that of \SystemNameO. }


\thanks{
\protect\\
Manuscript received July 3, 2020; revised August XX, 2020.}}
\markboth{Journal of \LaTeX\ Class Files,~Vol.~14, No.~8, August~2015}%
{Shell \MakeLowercase{\textit{et al.}}: Bare Demo of IEEEtran.cls for Computer Society Journals}

\IEEEtitleabstractindextext{%

\begin{abstract}
The lightweight semi-supervised learning (LSL) strategy provides an effective approach of conserving labeled samples and minimizing model inference costs. Prior research has effectively applied knowledge transfer learning and co-training regularization from large to small models in LSL. However, such training strategies are computationally intensive and prone to local optima, thereby increasing the difficulty of finding the optimal solution. This has prompted us to investigate the feasibility of integrating three low-cost scenarios for text mining tasks: limited labeled supervision, lightweight fine-tuning, and rapid-inference small models. We propose \SystemNameO, a novel framework for lightweight text mining that implements parameter-efficient learning with limited supervision. It employs online knowledge distillation to generate multiple small models and enhances their performance through mutual learning regularization. 
The entire process leverages parameter-efficient learning, reducing training costs and minimizing supervision requirements, ultimately yielding a lightweight model  for  downstream inference.
\end{abstract}

}

\maketitle
\IEEEdisplaynontitleabstractindextext
%
\IEEEpeerreviewmaketitle

\vspace{0.45in}
\IEEEraisesectionheading{\section{Introduction}\label{sec:introduction}}

\IEEEPARstart{S}{emi-supervised} learning (SSL)~\cite{reddy2018semi},\cite{wang2022usb} serves as a practical paradigm to enhance model generalization by utilizing scarce labeled data and abundant unlabeled data~\cite{lee2013pseudo},\cite{TarvainenV17},\cite{FurlanelloLTIA18},\cite{MiyatoMKI19},\cite{BerthelotCGPOR19},\cite{SohnBCZZRCKL20},\cite{fan2022revisiting},\cite{ZhangWHWWOS21},\cite{BerthelotRSCK22},\cite{ZhengYHWQX22}.  
Integrating self-supervised learning into lightweight descendants of pre-trained language models confronts two fundamental obstacles under scarce supervision: the paucity of labeled data constrains the acquisition of task-specific knowledge, while the absence of adequate regularization impairs generalization capacity.
Existing literature mitigates these challenges within lightweight semi-supervised learning (LSL).
\SystemNameA~\cite{JiangMLLDYW23} employs co-training to enlarge student generalization: heterogeneous views extracted from distinct teacher layers and augmented data jointly supervise the learner.
\SystemNameB injects adversarial perturbations that progressively intensify task difficulty, forcing the student to refine its decision boundaries and achieve self-improvement.
The former adopts offline distillation, the latter adopts online distillation, permitting real-time parameter updates and continuous adaptation.
Both frameworks compress large-scale teachers through knowledge distillation, couple this procedure with consistency regularization—co-training or mutual learning—and incorporate adversarial disturbances to regularize optimization, avoid poor local minima, and enhance the generalization capacity of lightweight student cohorts.

Contemporary pre-trained language models (PLMs) universally comprise decoder-only and encoder-only architectures. 
Decoder-only architectures, exemplified by GPT\cite{radford2018improving},\cite{radford2019language},\cite{brown2020language},\cite{openai2023gpt4},  LLaMA~\cite{touvron2023llama},\cite{touvron2023llama2},\cite{meta2024llama3},\cite{dubey2024llama31}, and DeepSeek\cite{deepseek2024lm},\cite{deepseek2024v3},\cite{deepseek2025r1}, constitute the dominant paradigm in generative AI and have driven  transformative advances in text generation.
Nonetheless, their parameter counts and computational overhead grow exponentially: GPT-1\cite{radford2018improving} contains 1.17\(\times 10^{8}\) parameters, whereas the recently released LLaMA 3.1~\cite{dubey2024llama31} scales up to 4.05\(\times 10^{11}\). 
This dramatic expansion imposes prohibitive constraints on compute and deployment budgets, thereby significantly hindering widespread replication and practical deployment. 
Encoder-only models—BERT~\cite{DevlinCLT19}, RoBERTa~\cite{liu2019roberta} and the recently released MBERT~\cite{WarnerCCWHTGBLA25}—produce high-dimensional dense embeddings that compress the input sequence into a low-dimensional continuous representation, thereby functioning as representation-centric models.

Recently, encoder-only models~\cite{WarnerCCWHTGBLA25,abs-2505-15696} epitomised by ModernBERT have re-entered the spotlight; ModernBERT embodies a substantial advance in encoder design, harnessing contemporary data scales and architectural refinements to establish new performance-efficiency frontiers across diverse Natural Language Processing applications.
Encoders remain popular because their modest inference demands enable large-scale document processing and rapid discriminative inference.
They deliver an attractive quality–size trade-off, positioning them as a pragmatic alternative to heavier encoder-decoder or decoder-only architectures when massive corpora must be handled efficiently.
Warner1 et al.,~\cite{WarnerCCWHTGBLA25} release MBERT\(_{\rm BASE}\) and MBERT\(_{\rm LARGE}\)\footnote{\url{https://github.com/AnswerDotAI/MBERT}}, which attain state-of-the-art overall performance across diverse downstream tasks relative to all existing encoder models\footnote{On GLUE, MBERT\(_{\rm BASE}\) surpasses other similarly-sized encoder models, and MBERT-large is second only to Deberta-v3-large~\cite{HeLGC21}.}.  
We adopt MBERT from FlexBERT\footnote{The codebase of FlexBERT builds upon MosaicBERT~\cite{PortesTHKVNSKF23} under the terms of its Apache 2.0 license.} as the backbone in our model.  Its framework facilitates straightforward experimentation and provides all intermediate training checkpoints, enabling it to serve directly as the teacher network in our model. Concurrently, our architecture adapts the canonical BERT framework to accommodate ultra-lightweight parameter budgets.
Motivated by the resurgence of encoder-only representation models, we contend that the systematic reintegration of state-of-the-art generative advances into the encoder-only regime constitutes a critical research frontier, particularly for low-resource scenarios where text classification and semantic matching must operate under stringent computational constraints.

Put briefly, we investigate how to effectively utilize compressed small models with limited labeled data and rapid fine-tuning to achieve strong generalization—balancing scarce labeled input, parameter-efficient training, and lightweight model inference within an end-to-end framework: 
\begin{itemize}[leftmargin=*]
\item \textbf{Scarce-label input}: extreme-label supervision incurs minimal annotation cost, making it feasible to scale to massive datasets with very few human-provided labels.
\item \textbf{Parameter-efficient training}: parameter-efficient learning reduces training overhead by updating only tiny weights, enabling rapid adaptation on consumer-grade hardware.
\item \textbf{Lightweight model inference}: the final model stays compact for efficient deployment on devices with strict memory and power constraints.
\end{itemize}
Yet the triple constraint—\textcolor{black}{\scshape{Label-scarce supervision}}, \textcolor{black}{\scshape{parameter-efficient training}}, \textcolor{black}{\scshape{lightweight model inference}}—rarely holds. Scarce labels supply inadequate supervision, impeding task-specific knowledge acquisition, whereas extreme compression removes generic regularization and compromises generalization.
To simultaneously resolve model compression, label scarcity, and training acceleration, we introduce \textbf{~\SystemNameO}\footnote{Code available at:\url{https://github.com/LiteSSLHub/NanoNet}}: a parameter-efficient framework that distils label-scarce supervision into lightweight encoder-only PLMs for rapid text-mining deployment.

We initially obtain a compact model through knowledge distillation, wherein a large-scale teacher network compresses its learned representations into a substantially smaller student network. To accommodate label scarcity, the student is subsequently optimized within a semi-supervised learning paradigm that leverages a minimal labeled subset together with with an extensive pool of unlabeled instances.
Furthermore, we instantiate an ensemble of lightweight student networks that engage in reciprocal instruction, thereby enabling collaborative optimization via mutual knowledge exchange\footnote{Ensembling low-bias student models and encouraging high diversity among their predictive distributions can jointly maximize the ensemble's posterior entropy, yielding better generalization bounds with inductive biases~\cite{ZhangXHL18},\cite{XieDHL020},\cite{LoveringJLP21},\cite{LeeKKCCH22},\cite{JiangMLLDYW23},\cite{MaoJLLLWLL25}.}. 
The student assimilates the distilled knowledge from the teacher alongside diversified peer knowledge, enabling stable convergence even under severe label constraints. 
The abundant unlabeled data provide complementary distributional cues, facilitating the acquisition of broadly transferable feature representations. Throughout the training trajectory, we exclusively adopt the BitFit strategy: only bias parameters are updated, while all remaining weights remain frozen. This yields parameter-efficient fine-tuning with minimal computational overhead.

At its core, \SystemNameO performs online distillation to compress the teacher, applies mutual learning to sculpt the optimization surface, and employs selective bias-term updates to accelerate training, thereby systematically expanding the generalization capacity of the lightweight student cohort. 
Comprehensive experiments on standard semi-supervised text-classification benchmarks  demonstrate that the lightweight instantiations of our \SystemNameO framework achieve superior performance under a label budget of merely 10/30/40/50 annotated samples.

The teacher-student backbone is instantiated with MBERT~\cite{WarnerCCWHTGBLA25}, whose alternating global and 128-token local attention mechanisms yield asymptotically lower computational complexity on long sequences and enable variable-length batch packing without padding overhead, thus beyond strictly satisfying the triple constraint, \SystemNameO yields additional computational savings.
Relative to the full 12-layer BERT and 24-layer MBERT backbones, the lightweight \SystemNameO variants simultaneously reduce trainable parameters, deployment size, and inference cost by orders of magnitude.


To sum up, the contributions of this paper are as follows:
\begin{itemize}[leftmargin=*]
\item We empirically reveal that existing SSL and LSL pipelines for encoder-only PLMs still struggle to simultaneously satisfy scarce-label input, parameter-efficient training, and lightweight model inference.
\item As a solution, we propose a new framework \SystemNameO, which orchestrates online knowledge distillation, semi-supervised learning, and parameter-efficient training into a unified framework.
\item We propose a reciprocal-instruction ensemble: the teacher distills knowledge into ultra-lightweight students, students mutually refine, and the framework performs LSL while updating only the bias terms, achieving parameter-efficient training under extreme-label supervision and yielding the final lightweight inference model. 
\item Extensive experiments on  text-classification benchmarks\footnote{\url{https://huggingface.co/datasets/LiteSSLHub/NanoNet/Textmining}, \url{https://huggingface.co/datasets/LiteSSLHub/NanoNet/Textmining-USB}} demonstrate the effectiveness of our framework: A two-layer distilled \SystemNameO\ student delivers comparative or superior accuracy relative to the leading baselines \SystemNameA\ and \SystemNameB\ under an identical sparse-label regime, while it reduces the trainable parameter count by at least \(0.9\times10^{3}\), and sustains inference latency comparable to that of the baselines.
\end{itemize}

\section{BACKGROUND AND PRELIMINARY STUDY}
\subsection{Consistency-based SSL with Label-Scarce Supervision}
Ensemble knowledge fusion enhances light semi-supervised learning (LSL) by aligning teacher–student predictions on unlabeled data through a consistency constraint~\cite{XieGDTH17}.  This strategy originates in network-noise regularization~\cite{SietsmaD91},\cite{GoodfellowSS14},\cite{MiyatoMKI19},\cite{KeWYRL19}: the teacher is an exponential moving average (EMA) of the student. 
VAT~\cite{MiyatoMKI19} and \(Pi\)-model~\cite{LaineA17} instantiate the teacher as the student itself (EMA coefficient zero), whereas the Temporal Model~\cite{LaineA17} additionally averages past predictions.
Mean Teacher~\cite{TarvainenV17} forms the teacher through an EMA of the student weights.
Ke et al.~\cite{KeWYRL19} demonstrate that this tight coupling increasingly constrains performance.
To relax it, Deep Co-Training~\cite{QiaoSZWY18} imposes consistency across independently initialized networks.
Dual-Student~\cite{KeWYRL19} removes the teacher entirely, and Jiang et al.~\cite{JiangMLLDYW23} augment this strategy with complementary data- and model-level noise, enabling fully independent model training.

Consistency-based SSL instantiates a Teacher–Student paradigm. Formally, it receives a semi-supervised dataset \(\mathcal{D}\), \(\mathcal{D}\!=\!\mathcal{S}\cup\mathcal{U}\).
\(\mathcal{S}\!=\!\left \{ (\hat{x},\hat{y})\right \}\) is labeled data and  
\(\mathcal{U}\!=\!\left \{ x^{*}\right \}\) is unlabeled data, and both students and teacher use all data identically.  
Let \(\Theta^{*}\) denote the weights of the teacher, and \(\Theta\)  denote the weights of the student. The consistency constraint is defined as:
\begin{equation}
\mathcal{L}_{\text{con}} = \mathbb{E}_{x^{*}\in \mathcal{D}}\,
\mathcal{R}\bigl(f(x^{*}+\eta, \Theta), \mathcal{T}_{x^{*}}, \Theta^{*}\bigr),
\label{eq:consistency}
\end{equation}
where $f(x^{*}+\eta)$ denotes the student prediction on noise-augmented input $x^{*}+\eta$.  $\mathcal{T}$ is the consistency target produced by the teacher, and $\mathcal{R}(\cdot)$ measures the distance between two vectors, usually chosen as the mean squared error (MSE) or the KL-divergence.

Previous works have proposed several strategies to generate $\mathcal{T}$.
Deep Co-Training, \SystemNameA, and \SystemNameB are consistency-based SSL frameworks that natively host cohorts of identical or near-identical students. Like cross-cohort consistency regularization lets every student refine its predictions against the others and lifts the whole cohort's performance.
Considering \(K\) networks \(\Theta_{1}\),...,\(\Theta_{i}\),...,\(\Theta_{K}\)(\(K\geq 2\)), the objective function for optimising all \(\Theta_{k}\), (\(1\!\leq \!k\leq \!K\)), becomes: 
\begin{equation}
\mathcal{L}_{\text{con}} = \mathbb{E}_{x^*\!\in\mathcal{D}}\;
\!\frac{1}{K\!-\!1}\!\sum_{i=1,i\neq k}^{K}\;
\mathcal{R}\!\left(f_{\Theta_k}(x^*\!\!+\eta, \Theta_k),\;
f_{\Theta_i}(x^*\!\!+\eta,\Theta_i)\right),
\label{eq:cohort}
\end{equation}
where \(\mathcal{R}\) is the cohort-level consensus target, typically measured by MSE or KL-divergence.
When extending the cohort to more than two networks, each student learns from the ensemble of the other peers, mimicking their collective output.

\subsection{Faster and Lighter SSL with Label-Scarce Supervision}
Recent years witness a surge of interest in faster and lighter  semi-supervised learning that coalesces around three complementary axes: (i) architecture-oriented distillation, (ii) cohort-wise co-training, and (iii) Online mutual regularization. 
(i) FLiText~\cite{LiuZFHL21} pioneers methodology (i) by pairing a frozen BERT `inspirer' with consistency regularization, enabling TextCNN and LSTM to reach 90.49\% IMDb accuracy with <1\% labels while keeping the student 30\(\times \) faster than the teacher. 
(ii) \SystemNameA~\cite{JiangMLLDYW23} embodies methodology (ii) and shows that two lightweight students can co-train effectively by exchanging diverse-augmented views without ever touching a heavy offline teacher, cutting computational cost by up to 7.6\(\times \) smaller compared with standard PLMs pipelines.  
(iii) \SystemNameB instantiates methodology (iii) and replaces \SystemNameA's offline distillation with on-the-fly mutual distillation, eliminating the need for pre-trained teachers.  Collectively, these efforts delineate a unified blueprint for rapid, lightweight SSL: aggressive compression through distillation or parameter sharing, and ensemble consistency driven by low-cost augmentation.

From an optimization perspective, faster-and-lighter SSL could simply to  combine supervised knowledge optimization of cross-entropy loss \(\mathcal{L}_{coe}\), unsupervised model compression of knowledge distillation loss \(\mathcal{L}_{kd}\) and consistency semi-supervised loss \(\mathcal{L}_{con}\):
\begin{equation}
\mathcal{L}_{\text{coe}}=\mathcal{L}_{T}\!\sum _{(\hat{x},\hat{y}) \in \mathcal{S}}f(\hat{x}, \Theta^{*}) + \sum _{k=1}^{K} \mathcal{L}_{S}^{k}\!\sum _{(\hat{x},\hat{y}) \in \mathcal{S}}f(\hat{x}, \Theta_{k}),
  \label{loss_multi}
\end{equation}
\begin{equation}
  \mathcal{L}_{\text{final}}=\mathcal{L}_{\text{coe}} + \sum _{k=1}^{K} \left(\mathcal{L}_{\text{kd}}^{k}\!+\!\mu(t,n)\cdot\lambda\cdot\mathcal{L}_{\text{con}}^{k}\right).
    \label{loss_multi}
\end{equation}

\begin{figure}[!t]
  \centering
  \begin{minipage}{.47\linewidth}
    \centering
    \includegraphics[width=\linewidth]{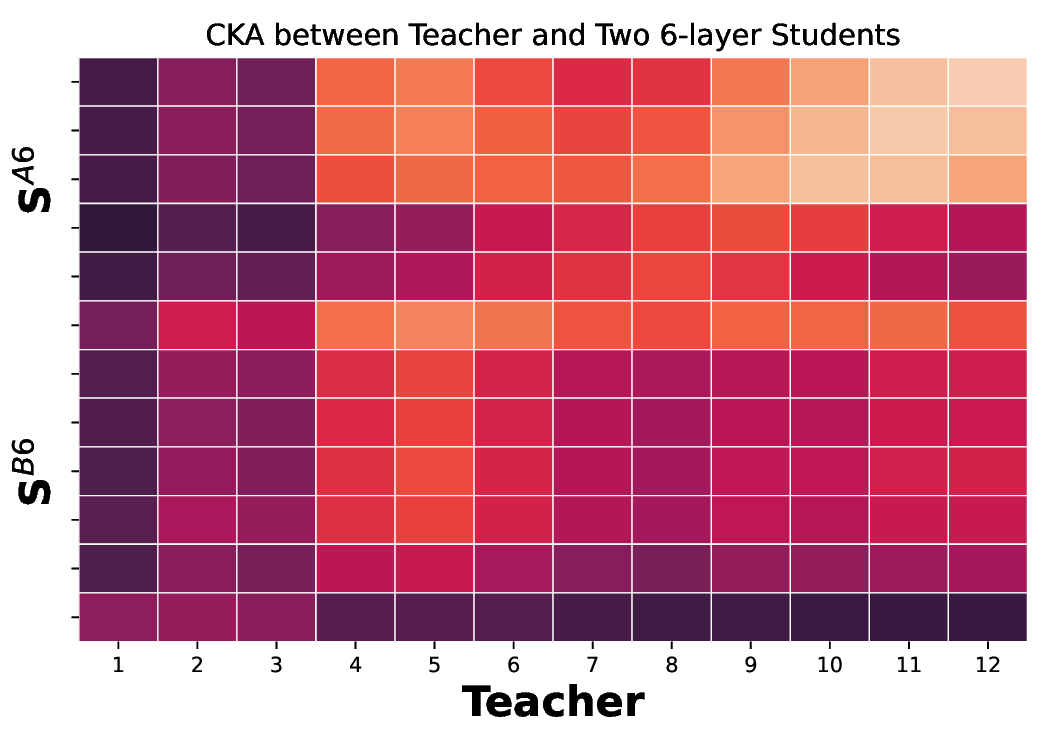}
    \centerline{\scriptsize (a) SingleStudent$^{\rm A6}$}
  \end{minipage}\hfil
  \begin{minipage}{.47\linewidth}
    \centering
    \includegraphics[width=\linewidth]{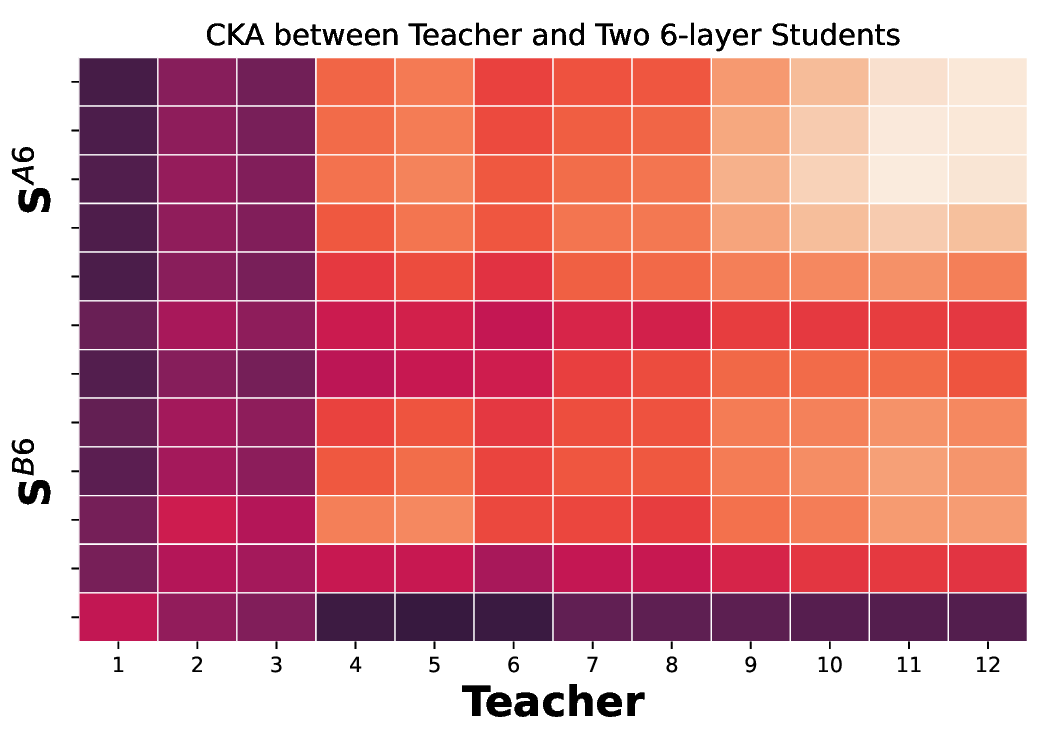}
    \centerline{\scriptsize (b) SingleStudent$^{\rm B6}$}
  \end{minipage}

  \vspace{4pt}   

  \begin{minipage}{.47\linewidth}
    \centering
    \includegraphics[width=\linewidth]{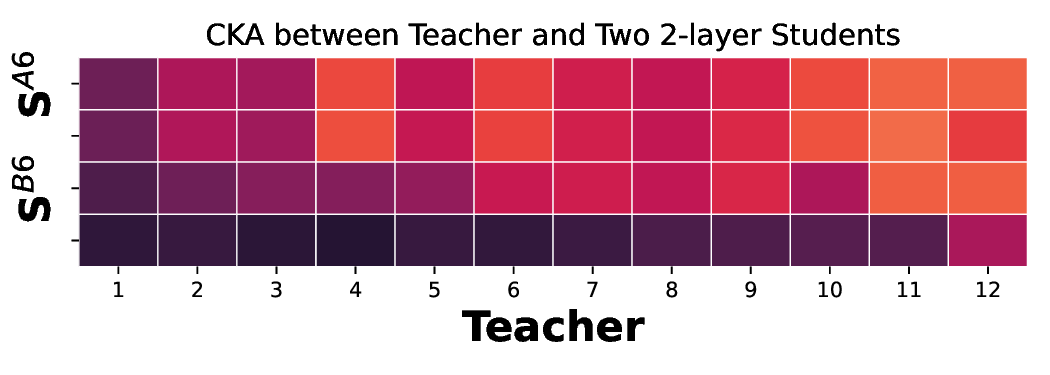}
    \centerline{\scriptsize (c) \SystemNameB $(\rm S^{A6})$}
  \end{minipage}\hfil
  \begin{minipage}{.47\linewidth}
    \centering
    \includegraphics[width=\linewidth]{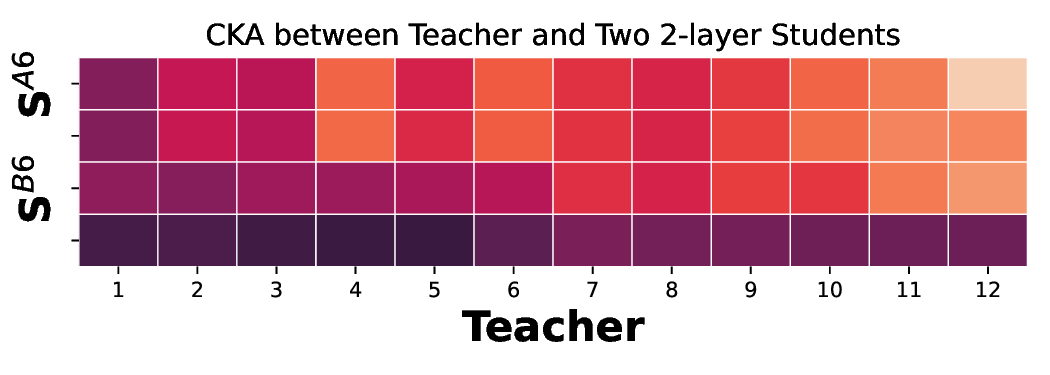}
    \centerline{\scriptsize (d) \SystemNameB $(\rm S^{B6})$}
  \end{minipage}

  \caption{The visualization of the Center Kernel Alignment (CKA~\cite{ZhuW21a}) scores of \SystemNameB in Subfigures (c) and (d), along with its ablation variant, SingleStudent, shown in Subfigures (a) and (b). All models are equipped with 6-layer BERT. The evaluation is conducted on text classification tasks using the AG News dataset, featuring 10 labeled data instances per class.}
  \label{heat_map}
\end{figure}

The objective reveals a dual knowledge path: a teacher stream that supplies consistency targets to every student, and a peer stream in which each student receives additional regularisation signals from its cohort. 
Figure~\ref{heat_map} visualises this effect for two students distilled from a 12-layer pretrained BERT\(_{\rm BASE}\). Panel (a) depicts plain online distillation that maps layers 1–6 to student 1 and layers 7–12 to student 2 without mutual exchange. Panel (b) augments the same setup with consistency-based mutual learning during semi-supervised fine-tuning. Panels (c) and (d) display the resulting two-layer student architectures.


The visualization results confirm that mutual learning enhances the Center Kernel Alignment (CKA) score between students and teachers, reducing the gap between them. This can be attributed to mutual learning indirectly extracting regularization information from each other's predictive capabilities, thereby enhancing the individual generalization abilities of the students.
While this approach is a typical example of Faster and Lighter SSL, it also introduces a new challenge: the teacher model must participate in parameter fine-tuning alongside the student models. 
Although the final deployment only requires a lightweight student model, the training process itself is computationally expensive. 
This raises the question of whether a more parameter-efficient fine-tuning strategy exists to reduce the computational overhead.

\begin{figure*}[!t]
  \centering
  \begin{minipage}{.553\linewidth}
    \centering
    \includegraphics[width=0.74\linewidth]{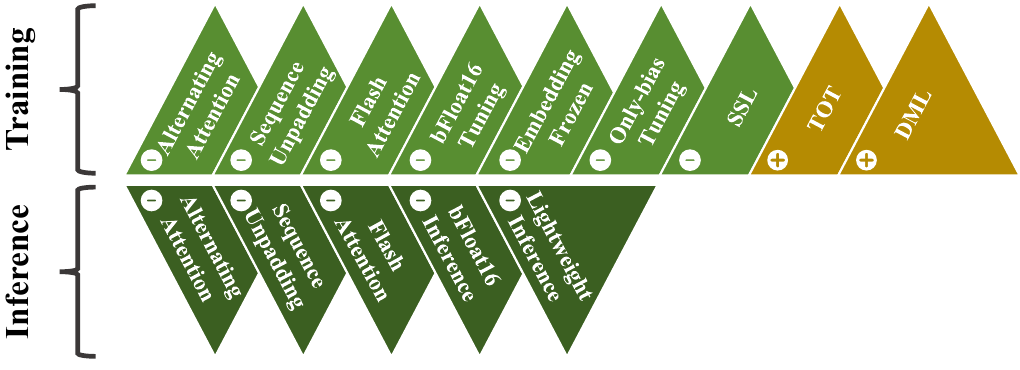}
    \centerline{(a) \SystemNameO}
  \end{minipage}\hfil
  \begin{minipage}{.22\linewidth}
    \centering
    \includegraphics[width=0.74\linewidth]{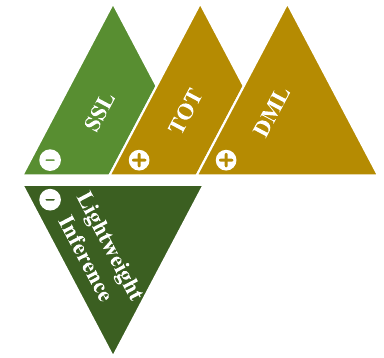}
    \centerline{(b) \SystemNameB}
  \end{minipage}\hfil
  \begin{minipage}{.22\linewidth}
    \centering
    \includegraphics[width=0.74\linewidth]{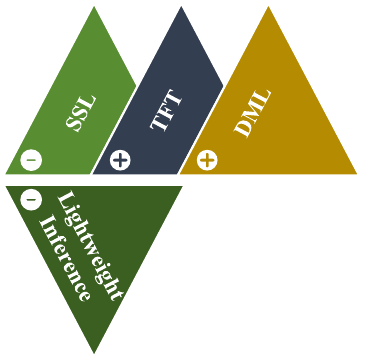}
    \centerline{(c) \SystemNameA}
  \end{minipage}
  \caption{The lightweight training modules and settings for the (a) \SystemNameO, (b) \SystemNameB and (c) \SystemNameA during the training and inference stages are as follows: a minus sign  represents the lightweight setting, a plus sign represents an additional module, SSL stands for Semi-Supervised Learning, the acronym TOT denotes `Teacher and Student training with online knowledge distillation,' whereas TFT signifies the offline variant. DML represents Deep Mutual Learning.}
  \label{comparing}
\end{figure*}

\subsection{Differences among Typical Lightweight SSL Models}
\SystemNameA\ and \SystemNameB\ pursue low-sample training for lightweight inference and a systematic comparison therefore proceeds.
All three models share training-architecture motifs and distillation strategies and the principal divergence is that \SystemNameO\ instantiates parameter-efficient training whereas the remaining two do not. 
As illustrated in Figure~\ref{comparing}, the models exhibit a fundamental similarity: \SystemNameA, \SystemNameB, and \SystemNameO each constitute a unified technical framework equipped with semi-supervised learning (SSL)  and deep mutual-learning (DML) that accelerates and streamlines consistency-based SSL. 
Specifically, they integrate SSL paradigm during the training phase and a lightweight inference mechanism during the deployment phase. With respect to semi-supervised optimization, the models converge on a common objective with DML: they enhance generalization performance by orchestrating complementary learning across multiple peer networks, thereby increasing the posterior entropy of every student network and fostering a shared representational experience.
Nevertheless, \SystemNameB introduces a markedly superior formulation of complementary learning that is distinguished by an online distillation mechanism, the complete absence of external dependencies, and a progressive training schedule. Conversely, \SystemNameO pursues architectural minimalism: it dispenses with the external data-augmentation submodule that characterizes \SystemNameA and removes  the adversarial regularization term that underpins \SystemNameB.

\subsubsection{\textbf{Distinct Origins and Methodologies in Offline vs. Online Distillation Methods}} 
\SystemNameA adopts an offline-distillation paradigm (termed as TFT, as illustrated in Figure~\ref{comparing}) that necessitates a two-stage pipeline: it first produces a generic student model and subsequently fine-tunes this model for the target task.
In contrast, \SystemNameB  employs online distillation (termed as TOT, as illustrated in Figure~\ref{comparing}), where supervised learning and unsupervised distillation for both the teacher and student occur simultaneously. 
This synchronous regime yields a fluid, continuous approximation of the teacher's decision boundaries: generic representations are consolidated in the earliest training phases, whereas task-specialized knowledge is progressively refined in subsequent stages.
This consistent emulation allows the teacher to guide the optimization paths of all students, rather than relying solely on peer learning without external guidance. By emphasizing external guidance, \SystemNameB enhances mutual learning between two divergent students, ultimately strengthening their collaborative learning process. \SystemNameO, conversely, reverts to offline distillation: it transfers the teacher's pre-consolidated parameters to the students, thereby reducing the fine-tuning stage to a lightweight optimization problem—implemented through approximately seven parameter-efficient techniques illustrated in Figure~\ref{comparing}—and accelerates convergence toward a global optimum. \SystemNameO adopts offline knowledge distillation, because this strategy obviates the prerequisite of pre-training the teacher model on the target data, a condition that online distillation imposes.

\subsubsection{\textbf{Distinct Origins and Methodologies in New Learning Procedures}} 
\SystemNameO employs advanced parameter-efficient  training methods that ensure the training process requires only a small amount of labeled data and a limited number of training  parameters, and that the final inference model is also small. As shown, most modules in the training and inference stages are not present in \SystemNameA and \SystemNameB. \SystemNameA conducts co-training by processing labeled and unlabeled data together in a single batch.
In contrast, \SystemNameB performs supervised learning  sequentially followed by unsupervised knowledge distillation. This approach encourages the teacher model to actively participate in every single optimization step, thereby mitigating the impact of the scale gap between the teacher and student models on distillation performance. Similar discussions can be found in methodologies such as TAKD~\cite{MirzadehFLLMG20} and BANs~\cite{FurlanelloLTIA18}. Furthermore, \SystemNameB  incorporates curriculum adversarial training (CAT) to  progressively increase learning complexity. This enables \SystemNameB  to implement an iterative learning approach, facilitating continuous self-improvement of the lightweight model.

\subsection{Selective Parameter-efficient Fine-tuning}
Parameter-efficient fine-tuning (PEFT)~\cite{abs-2404-13506} strikes  a balance between accuracy 
and efficiency by selectively updating a subset of model parameters. 
These methods have the potential to significantly reduce the computational costs and memory usage. Three major additive finetuning algorithms are normally used: (i) additive fine-tuning, (ii) reparameterized fine-tuning, and (iii) selective fune-tuning. 
Additive fine-tuning keeps the pre-trained backbone frozen and trains only lightweight task specific modules, such as adapters\cite{HoulsbyGJMLGAG19},\cite{HeZMBN22}, soft prompts\cite{LiL20},\cite{0006ABZ23},\cite{abs-2110-07602},\cite{ZhangTX00H23},\cite{LesterAC21},\cite{MaZRWWWQ022}, thereby reducing  the number of trainable parameters.
Reparametrization-based fine-tuning recasts the parameter update as an equivalent, low-rank transformation. Methods such as LoRA~\cite{HuSWALWWC22} and its extensions~\cite{ZhangQSX24,HayouG024,WuHW24} inject trainable low-rank matrices during optimization and subsequently merge these matrices with the frozen weights for inference, preserving the original architecture while constraining the parameters.
Selective fine-tuning introduces no additional parameters; instead, it designates a sparse, predetermined subset of the backbone weights as trainable and freezes the remainder. In contrast to additive approaches, which enlarge the effective model size, selective PEFT reuses existing parameters. This paradigm encompasses both unstructured schemes—U-Diff pruning~\cite{GuoRK20}, U-BitFit~\cite{LawtonKTGS23}, FishMask~\cite{SungNR21}—and structured schemes such as S-Diff pruning~\cite{GuoRK20}, S-BitFit~\cite{LawtonKTGS23}, SPT~\cite{He0ZTZ23}, BitFit~\cite{ZakenGR22} and its variants~\cite{LawtonKTGS23}. Unstructured masks, however, produce irregular sparsity patterns that degrade hardware utilization. Structured masks impose regular, hardware-amenable patterns, simultaneously enhancing computational and throughput efficiency; consequently, structured selective PEFT aligns seamlessly with our framework, whose lightweight mandate precludes the introduction of any supplementary network modules.


\section{Methodology of \SystemNameO}
\label{sec:Methodology_of_NANONET}
This section presents our novel framework for lightweight semi-supervised learning, which achieves parameter-efficient learning with limited supervision. Our \SystemNameO framework incorporates sequential unpadding modeling and selective parameter fine-tuning, both aimed at enhancing training efficiency. These components are detailed in Sections~\ref{s_u_m} and~\ref{s_p_f}, respectively.
Furthermore, we introduce an offline  knowledge distillation framework that compresses large models and optimizes small models during semi-supervised learning. Specifically, a limited amount of labeled  data is used for supervised learning (Section~\ref{s_m_o}), while a large amount of unlabeled data supports consistency-based regularization (Section~\ref{u_c_c}). This design enables our \SystemNameO framework to generate lightweight models with minimal supervision.

\begin{figure*}[tbp]
\centerline{\includegraphics[width=1.0\textwidth]{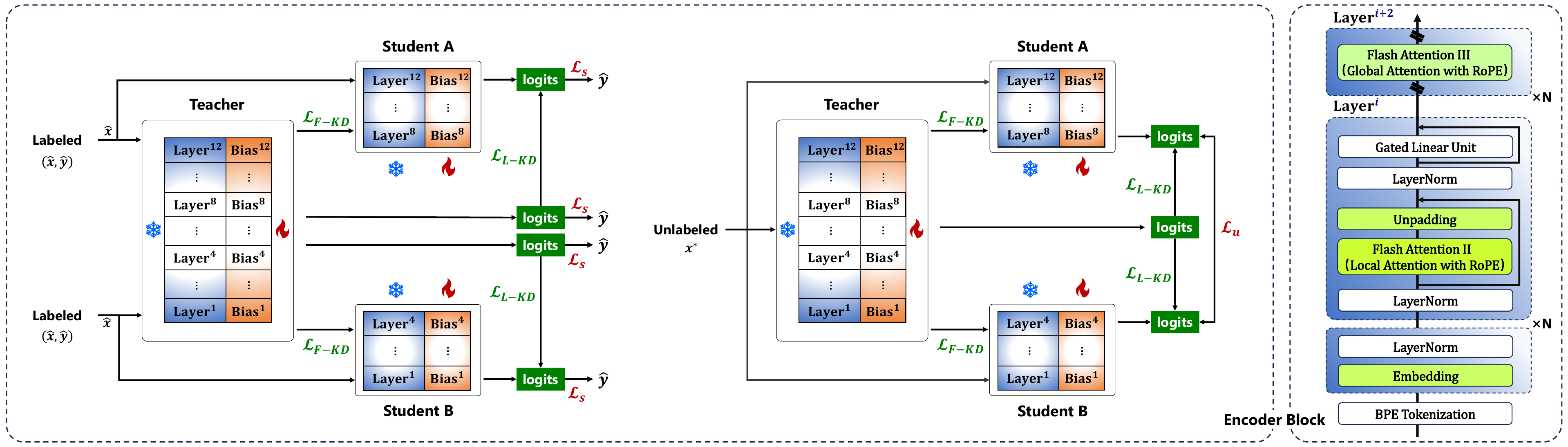}}
    \caption{(LEFT) Framework overview of the \SystemNameO framework. The Teacher model heuristically delegates distinct layers to distinct student models and throughout training only bias parameters are updated while all remaining weights stay frozen.
     (RIGHT) Schematic of Encoder Block. Following ModernBERT, unpadding is fused into a single Flash-Attention kernel by supplying the sequence-boundary indices obtained from tokenization directly as inputs to the attention computation. Global attention is applied at fixed sparse intervals with a high-capacity RoPE (Rotary Position Embedding), while the remaining layers employ local sliding-window attention with a correspondingly reduced RoPE.
}
    \label{framework}
\end{figure*}

We illustrate the dual-student \SystemNameO process for training two lightweight cohorts (see Figure~\ref{framework}). 
The framework is outlined as follows:
\begin{itemize}[leftmargin=*]
\item \SystemNameO integrates supervised learning for knowledge optimization and unsupervised distillation for model compression. It builds upon the concept of offline distillation from \SystemNameB , enabling multiple student models to be trained simultaneously. This approach allows the small models to leverage large amounts of unlabeled data, thereby enhancing  their performance through semi-supervised learning.
\item \SystemNameO employs mutual learning to train student cohorts, where each student mimics the output logits of other peers within a collaborative setup. This mutual imitation facilitates knowledge exchange among students, allowing them to acquire diverse perspectives and enhancing their collective generalization ability.
\item \SystemNameO optimizes efficiency and reduces fine-tuning costs via (at minimum)  three techniques: Unpadding Input Modeling removes padding tokens to cut computational waste, Alternating Attention trims parameters in forward fine-tuning passes, and Selective Parameters Fine-tuning updates only bias terms during both forward and backward fine-tuning passes to minimize computational load.
\end{itemize}
Below, we provide a detailed introduction.

\subsection{Sequential Unpadding Modeling}
\label{s_u_m}
Unpadding input modeling, following MBERT~\cite{WarnerCCWHTGBLA25}, MosaicBERT~\cite{PortesTHKVNSKF23}, leverages unpadding techniques~\cite{ZhangZLXDTLYXHZ24},\cite{abs-2208-08124} throughout both training and inference, thereby eradicating computational redundancies induced by padding tokens.
Conventional encoder-only models (such as BERT, RoBERTa)~\cite{DevlinCLT19},\cite{liu2019roberta} pad sequences to uniform lengths, wasting resources on semantically irrelevant tokens. Unpadding strips the filler tokens, concatenates every sequence into one contiguous stream, and feeds the resulting tensor through the encoder once; because no layer ever re-pads or re-unpads, all redundant computation and memory traffic are eliminated.

\subsection{Offline Knowledge Distillation}
\label{o_k_d}
Formally, we are provided with a semi-supervised dataset $\mathcal{D}$, $\mathcal{D} = \mathcal{S} \cup \mathcal{U}$. $\mathcal{S} = \{(\hat{x}, \hat{y})\}$ is labeled data, where $(\hat{x}, \hat{y})$ will be used for two kinds of students identically. $\mathcal{U} = \{x^*\}$ is unlabeled data, and two copies are made for two kinds of students identically. 
Offline Knowledge Distillation in \SystemNameO constitutes a dynamic training paradigm that leverages the unlabeled set \(\mathcal{U}\) to enable two compact student models to acquire knowledge from a pre-trained, larger teacher model while permitting both models to be updated simultaneously.
This strategy mitigates the training-inference discrepancy inherent in online distillation by equipping the student models to assimilate the teacher's behavioral characteristics with heightened fidelity.
The methodology confers multiple advantages: it accelerates convergence, facilitates the acquisition of difficult instances through the teacher's softened predictions, and diminishes dependence on extensive labeled corpora.
Furthermore, the \SystemNameO  retains the option to freeze the embedding parameters throughout training, thereby curtailing the student models'  parametric reliance on heavyweight embeddings and enabling markedly faster training.

In our framework, the teacher model is characterized as a deep bidirectional encoder, while the student model is designed as a lightweight architecture with a reduced number of layers. 
To streamline notation, we denote a model comprising \(k\) Transformer layers as Transformer\(k\).  
Within each layer, the Transformer employs multiple self-attention heads to integrate the output vectors from the preceding layer. For the \(l\)-th Transformer layer, the output of \(a\)-th self-attention head, denoted as \(\text{\textbf{AO}}^{a,l}\) is calculated as follows, where \(A_{h}\)  signifies the total number of self-attention heads and \(a\in [1, A_{h}]\):

\begin{equation}
\text{\textbf{Q}}^{a,l}(\text{\textbf{H}},\text{\textbf{W}},\text{\textbf{\textcolor{blue}{b}}}) = \text{\textbf{W}}_q^{a,l}\, \text{\textbf{H}}^{l-1} + \text{\textbf{\textcolor{blue}{b}}}_q^{a,l},
\label{eq5}
\end{equation}
\begin{equation}
\text{\textbf{K}}^{a,l}(\text{\textbf{H}},\text{\textbf{W}},\text{\textbf{\textcolor{blue}{b}}}) = \text{\textbf{W}}_k^{a,l}\, \text{\textbf{H}}^{l-1} + \text{\textbf{\textcolor{blue}{b}}}_k^{a,l},
\end{equation}
\begin{equation}
\text{\textbf{V}}^{a,l}(\text{\textbf{H}},\text{\textbf{W}},\text{\textbf{\textcolor{blue}{b}}}) = \text{\textbf{W}}_v^{a,l}\, \text{\textbf{H}}^{l-1} + \text{\textbf{\textcolor{blue}{b}}}_v^{a,l}. 
\end{equation}

Here, \(\text{\textbf{H}}\) represents the output of the preceding encoder layer. Each self-attention layer is composed of M distinct head networks. These head networks transform the input features, and their outputs are subsequently merged through an attention mechanism that does not introduce any additional parameters.

\begin{equation}
\text{\textbf{A}}^{a,l}= \text{softmax}\left(\frac{\text{\textbf{Q}}^{a,l} \left(\text{\textbf{K}}^{a,l}\right)^\top}{\sqrt{d_k}}\right) \text{\textbf{V}}^{a,l},
\end{equation}
\begin{equation}
\text{\textbf{AO}}^{a,l}=\text{\textbf{A}}^{a,l}\text{\textbf{V}}^{a,l}
\end{equation}

where the previous layer \(\text{\textbf{H}}^{l-1} \in \mathbb{R}^{L\times d_{h}}\) is linearly
projected into three components: queries, keys and values. This projection is achieved using parameter matrices  \(\text{\textbf{W}}_q^{a,l}\), \(\text{\textbf{W}}_k^{a,l}\), \(\text{\textbf{W}}_v^{a,l} \in \mathbb{R}^{d_{h}\times d_{k}}\), respectively. The attention distribution \(\text{\textbf{A}}^{a,l} \in \mathbb{R}^{L\times L}\) is calculated through the scaled dot-product of the queries and keys. 
Multi-Head Attention (MHA) concatenates specialised-head outputs to form the joint representation.

\begin{equation}
\text{MHA}^l(\text{\textbf{H}},\text{\textbf{W}},\text{\textbf{\textcolor{blue}{b}}}) = \text{concat}\left(\text{\textbf{A}}^{1,l}, \ldots, \text{\textbf{A}}^{A_h,l}\right) \text{\textbf{W}}_o^l + \text{\textbf{\textcolor{blue}{b}}}_o^l, 
\end{equation}

\subsubsection{\textbf{Self-Attention Distribution Transfer}}
Some works have shown that self-attention distributions of pre-trained LMs capture a rich hierarchy of linguistic information~\cite{JawaharSS19,WangW0B0020}.  
We utilize the self-attention distributions to assist 
the training of the student. Specifically, we minimize the KL-divergence between the self-attention distributions of the teacher and the student: 

\begin{equation}
\mathcal{L}_{\text{attn}_{T\leftrightarrow S}} = \frac{1}{A_h \, L} \sum_{a=1}^{A_h} \sum_{l=1}^{L} \text{MSE}(\text{MHA}_T^{N,a,l}\, , \, \text{MHA}_S^{M,a,l}),
\end{equation}

where, \(L\) denotes the sequence length, while \( A_h \) signifies the number of attention heads. Additionally, \(N\) and \(N\) indicate the number of layers within the teacher and student models, respectively. The attention distributions for the teacher and student at the \( l \)-th Transformer layer are represented by \(\text{MHA}_T^{N,a,l} \) and \(\text{MHA}_S^{M,a,l} \). These distributions facilitate the layer-by-layer transfer of the teacher's knowledge to the student. The attention distributions are determined through a scaled dot-product operation involving queries and keys.

\subsubsection{\textbf{Hidden-States Distribution Transfer}}
Beyond attention distillation, we also distil the intermediate layers as following  procedures.

\begin{equation}
m^l\!(\text{\textbf{H}},\!\text{\textbf{W}},\!\text{\textbf{\textcolor{blue}{b}}}) =\text{\textbf{W}}_{p}^l G_{\jmath} \!\left(\text{\textbf{W}}_{f}^l\! \left(\text{MHA}^l\!(\text{\textbf{H}},\!\text{\textbf{W}},\!\text{\textbf{\textcolor{blue}{b}}}) + \text{\textbf{H}}^{l-1}\!(\text{\textbf{H}},\!\text{\textbf{W}},\!\text{\textbf{\textcolor{blue}{b}}})\!\right)\right)\!, 
\end{equation}

where \(G_{\jmath}\) represents the GeGLU activation function.
\begin{equation}
\text{\textbf{H}}^l(\text{\textbf{H}},\!\text{\textbf{W}},\!\text{\textbf{\textcolor{blue}{b}}}) \!=\! \text{\textbf{H}}^{l-1}(\text{\textbf{H}},\!\text{\textbf{W}},\!\text{\textbf{\textcolor{blue}{b}}}) + \text{MHA}^l(\text{\textbf{H}},\!\text{\textbf{W}},\!\text{\textbf{\textcolor{blue}{b}}}) + m^l(\text{\textbf{H}},\!\text{\textbf{W}},\!\text{\textbf{\textcolor{blue}{b}}}),
\end{equation}

The projection layer weight matrix, denoted as \( \mathbf{W}_{\text{p}}^l \), is utilized to adjust the feature representations that have been processed by a nonlinear activation function, specifically the sigmoid function \( \sigma \). Meanwhile, the fully connected layer weight matrix, represented by \( \mathbf{W}_{\text{f}}^l\), primarily serves to perform a linear transformation on the outputs from the multi-head attention layer as well as the outputs from the previous layer.

\begin{equation}
\mathcal{L}_{\text{hidden}_{T\leftrightarrow S}} = \frac{1}{L} \sum_{l=1}^{L} \text{MSE}(\text{\textbf{H}}_T^l\, , \, \text{\textbf{W}}_{h}\text{\textbf{H}}_S^l),
\end{equation}
The matrix \( \mathbf{W}_h \in \mathbb{R}^{d_{h_S} \times d_{h_T}} \) represents a learnable linear transformation that maps the hidden states of the student network into the same feature space as those of the teacher network. The scalar values \( d_{h_T} \) and \( d_{h_S} \) denote the hidden sizes of the teacher and student models, respectively.

\subsubsection{\textbf{Output-Logits Distribution Transfer}}
Beyond mimicking the behavior of intermediate layers, we also employ knowledge distillation to align the predictions\footnote{By learning the logits of the teacher model, the student model acquires an understanding of the teacher's confidence in its predictions. Logits are generally smoother than the probability distributions produced by the softmax function, which aids the student model in achieving more stable gradients during training~\cite{HintonVD15,JiangMLLDYW23,SunR00C24,ZhangL0M24,MaoJLLLWLL25}.} of the teacher model with those of the student model. We achieve this by penalizing the soft cross-entropy loss computed when comparing the student network's logits, denoted as \(z^{x^{*}}_S(\text{\textbf{H}},\text{\textbf{W}},\text{\textbf{b}})\) against the teacher's logits \(z^{x^{*}}_T(\text{\textbf{H}},\text{\textbf{W}},\text{\textbf{b}})\):

\begin{equation}
z^{x^{*}}_T(\text{\textbf{H}},\text{\textbf{W}},\text{\textbf{b}}) = \text{\textbf{W}}_{logit} \, \text{\textbf{H}}_T^L(\text{\textbf{H}},\text{\textbf{W}},\text{\textbf{\textcolor{blue}{b}}}) + \text{\textbf{\textcolor{blue}{b}}}_{logit},
\end{equation}

\begin{equation}
z^{x^{*}}_S(\text{\textbf{H}},\text{\textbf{W}},\text{\textbf{\textcolor{blue}{b}}}) = \text{\textbf{W}}_{logit} \, \text{\textbf{H}}_S^L(\text{\textbf{H}},\text{\textbf{W}},\text{\textbf{\textcolor{blue}{b}}}) + \text{\textbf{\textcolor{blue}{b}}}_{logit},
\end{equation}

\begin{equation}
\mathcal{L}_{logit_{T\leftrightarrow S}} = \text{MSE}(z^{x^{*}}_T(\text{\textbf{H}},\text{\textbf{W}},\text{\textbf{\textcolor{blue}{b}}})  \, , \, z^{x^{*}}_S(\text{\textbf{H}},\text{\textbf{W}},\text{\textbf{\textcolor{blue}{b}}})/t), 
\end{equation}

where \(t\) means the temperature value. In our experiment, we find that \(t=1\) performs well.

\subsection{Unsupervised Cohort Co-training}
\label{u_c_c}
Drawing on the approaches of \SystemNameA and \SystemNameB, we improve the collaborative optimization process through the application of Deep Mutual Learning (DML)~\cite{ZhangXHL18} among student models. Although both  \SystemNameA and \SystemNameB are capable of incorporating numerous students into their mutual learning frameworks without any issues,  our priority is to reduce training expenses. As a result, we have chosen a dual-student learning strategy that includes only two student networks. The consistency cost for the unlabeled data \(x^{*}\)  is determined based on the output logits from these two student models:

\begin{equation}
\mathcal{L}_{u_{S_{A\to B}}} = \text{MSE}(z^{x^{*}}_{S_{A}}(\text{\textbf{H}},\text{\textbf{W}},\text{\textbf{\textcolor{blue}{b}}}), z^{x^{*}}_{S_{B}}(\text{\textbf{H}},\text{\textbf{W}},\text{\textbf{\textcolor{blue}{b}}})),
\end{equation}

\begin{equation}
\mathcal{L}_{u_{S_{B\to A}}} = \text{MSE}(z^{x^{*}}_{S_{B}}(\text{\textbf{H}},\text{\textbf{W}},\text{\textbf{\textcolor{blue}{b}}}), z^{x^{*}}_{S_{A}}(\text{\textbf{H}},\text{\textbf{W}},\text{\textbf{\textcolor{blue}{b}}})).
\end{equation}

DML operates  exclusively to the unsupervised partition, where the mean-squared error (MSE) between the logits of two peer students constitutes the training signal. Gradients are disabled for the counterpart branch via the detach operator, so each student (we employ the minimal set of two) is updated solely from its own distillation loss.

We exclude the multi-student setting from our experimental analysis. While prior studies (such as \SystemNameA and \SystemNameB) show that generalization improves as peer numbers grow—suggesting that a larger pool of student perturbations complement one another and boost the inference performance —we omit this strategy because multiplying students linearly increases training overhead, contradicting our core motivation of parameter efficiency learning.

\subsection{Supervised Model Optimization}
\label{s_m_o}
In supervised learning, we employ the Cross-Entropy (CE) loss function to optimize the parameters of both the student and teacher models and These models are trained using labeled data pairs $(\hat{x}, \hat{y})$ extracted from the dataset \( \mathcal{S} \). 
\begin{equation}
\mathcal{L}_{S_{A}} = CE(\text{softmax}(z^{\hat{x}}_S(\text{\textbf{H}},\text{\textbf{W}},\text{\textbf{\textcolor{blue}{b}}}), \hat{y})),
\end{equation}
\begin{equation}
\mathcal{L}_{S_{B}} = CE(\text{softmax}(z^{\hat{x}}_S(\text{\textbf{H}},\text{\textbf{W}},\text{\textbf{\textcolor{blue}{b}}}), \hat{y})),
\end{equation}
\begin{equation}
\mathcal{L}_{T} = CE(\text{softmax}(z^{\hat{x}}_T(\text{\textbf{H}},\text{\textbf{W}},\text{\textbf{\textcolor{blue}{b}}}), \hat{y})).
\label{eq22}
\end{equation}

\begin{table*}[t]
  \centering
  \caption{Dataset statistics and dataset split for semi-supervised text classification tasks, in which `\(\times\)' means the number of data per class. `\(-\)' means to subtract the quantity of data.}
  \renewcommand\arraystretch{1.1}
  \setlength{\tabcolsep}{4.4mm}{
  \begin{tabular}{|lccccrr|}
    \specialrule{1pt}{0pt}{0pt}
  \textbf{Dataset}       & \textbf{Label Type}               & \multicolumn{1}{c}{\textbf{Classes}} & \multicolumn{1}{c}{\textbf{Labeled}} & \multicolumn{1}{c}{\textbf{Unlabeled}}                                       & \multicolumn{1}{c}{\textbf{Dev}} & \multicolumn{1}{c|}{\textbf{Test}} \\ 
  \hline
  Agnews  in LiteSSLHub      & News Topic               & 4                          
  & \(\times 10/\times 30/\times 200\)                 & 20,000                                                                 & 8,000                     & 7,600                      \\ 
  Yahoo!Answer in LiteSSLHub   & Q\&A Topic                 & 10                         & \(\times 10/\times 30/\times 200\)                 & 50,000                                                                 & 20,000                    & 59,727                     \\ 
  DBpedia  in LiteSSLHub      & Wikipedia   Topic        & 14                         & \(\times 10/\times 30/\times 200\)                 & 70,000                                                                 & 28,000                    & 70,000                     \\
  Agnews  in USB      & News Topic               & 4                          
  & \(\times 40/\times 200\)                 & 99,800                                                                & 10,000                     & 7,600                      \\ 
  Amzn Review in USB       & Product Review Topic        & 5                         & \(\times 50/\times 200\)                 & 249,000                                                                 & 25,000                    & 65,000                     \\ 
   Yahoo!Answer in USB   & Q\&A Topic                 & 10                         & \(\times  500/\times 2000\)                 & 498,000                                                  & 50,000                    & 60000                     \\ 
   Yelp Review in USB      & Business Review Topic        & 5                         & \(\times 50/\times 200\)                 & 249,000                                                                 & 25,000                    & 50,000                     \\ 
   \specialrule{1pt}{0pt}{0pt}
  \end{tabular}
  }
  \label{statistics} 
\end{table*}

\begin{table}[tbp]
  \centering
  \caption{Definitions of Key Notations for Model Architecture and Training Parameters. $\mathcal{L}({\tiny M_{\text{Tea}}})$ as the number of layers in the teacher model within a framework, or the model's own layers if there is no teacher-student structure.   $\mathcal{L}({\tiny M_{\text{Stu}}})$ as the number of layers in the student model within a framework, or the model's own layers if there is no teacher-student structure.  $\mathcal{P}({\tiny M_{\text{Tra}}})$ as the parameters involved in the model training process, specifically for a teacher-student model structure. $\mathcal{P}({\tiny M_{\text{Inf}}})$ as the parameters of the final inference model, with the selection based on better performance on the validation set if there are two student models. `w/o E' refers to model training without tuning the embeddings. 
  N(ms) as the time cost for inferring a single sample, measured in seconds. 
  }
    \renewcommand\arraystretch{1.4}
  \setlength{\tabcolsep}{.25mm}{
    \begin{tabular}{|l|c|c|r|r|r|}
      \specialrule{1pt}{0pt}{0pt}
    \textbf{Models} & \multicolumn{1}{c|}{$\mathcal{L}({\tiny M_{\text{Tra}}})$} & \multicolumn{1}{c|}{$\mathcal{L}({\tiny M_{\text{Inf}}})$} & \multicolumn{1}{c|}{$\mathcal{P}({\tiny M_{\text{Tra}}})$} & \multicolumn{1}{c|}{$\mathcal{P}({\tiny M_{\text{Inf}}})$} & N(ms) \\
    \hline
    MBERT\(_{\rm LARGE}\)  & 28    & / &  396.18 &  396.83 &  13.53 \\
    MBERT\(_{\rm LARGE}\) w. BitFit & 28    & / & 51.93 &  396.18 &  13.53 \\
     BERT\(_{\rm BASE}\)  & 12    & / &  109.48 &  109.48 &  12.94 \\
    UDA   & 12    & / &  109.48 &  109.48 &  12.94 \\
    BitFit (BERT\(_{\rm LARGE}\)) & 24    & / & 26.81 &  335.14 &  13.01 \\
    TinyBERT\(^{4}\)   & 12    & 4     &  14.35 &  14.35 &  2.86 \\
    \SystemNameA (S\(^{\rm x2}\)) & 12    & 2     &  17.80 &  8.9 &  1.72 \\
    \SystemNameB (S\(^{\rm x2}\)) & 12    & 2     &  127.28 &  8.9 &  1.72 \\
    \textcolor{blue}{\SystemNameO} {\scriptsize (S\(^{\rm MBERT-x13}\)) w/o} E & 28    & 13    &  \textcolor{blue}{\textbf{0.69}} &  212.14 &  \textcolor{blue}{\textbf{6.27}} \\
    \textcolor{blue}{\SystemNameO} {\scriptsize (S\(^{\rm MBERT-x4}\)) w/o} E & 28    & 4     & \textcolor{blue}{\textbf{0.46}} &  101.71 & \textcolor{blue}{\textbf{3.14}} \\
     \textcolor{blue}{\SystemNameO} {\scriptsize (S\(^{\rm BERT-x6}\)) w/o} E & 12    & 6    &  \textcolor{blue}{\textbf{0.21}} &  \textcolor{blue}{\textbf{66.96}} &  \textcolor{blue}{\textbf{3.23}} \\
      \textcolor{blue}{\SystemNameO} {\scriptsize (S\(^{\rm BERT-x4}\)) w/o} E & 12    & 4     &  \textcolor{blue}{\textbf{0.18}} &  \textcolor{blue}{\textbf{52.78}} &  \textcolor{blue}{\textbf{2.67}} \\
    \textcolor{blue}{\SystemNameO} {\scriptsize (S\(^{\rm BERT-x2}\)) w/o} E & 12    & 2     &  \textcolor{blue}{\textbf{0.14}} &  \textcolor{blue}{\textbf{38.61}} &  \textcolor{blue}{\textbf{2.18}} \\
    \specialrule{1pt}{0pt}{0pt}
    \end{tabular}%
  }
  \label{addlabel-parameter}%
\end{table}%

\subsection{Selective Parameters Fine-tuning}
\label{s_p_f}
We implement additional efficiency-focused enhancements through architectural and implementation advancements as shown in Figure~\ref{comparing}. 
During the training process, we employ SSL approach, utilizing only a small amount of labeled data and we leverage unsupervised data for knowledge distillation to produce a compact student model. This student model is then used as the inference model, which reduces the parameter count of the inference model. 
To further minimize training costs, we implement a variety of parameter-efficient fine-tuning methods, including \textit{Alternating Attention}, \textit{Sequence Unpadding}, \textit{Flash Attention}, and \textit{bFloat16 Tuning}. Additionally, we freeze the embedding parameters, referred to as \textit{Embedding Frozen}, which removes 51.58 million trainable parameters and thereby achieves a substantial reduction in training cost.  
Our method offers a substantial reduction in computational overhead compared to these methods of \SystemNameA and \SystemNameB, with noticeable lightweight operations during both the training and inference phases.

\subsubsection{\textbf{Bias-Term Fine-Tuning}}
Moreover, the salient feature of our \SystemNameO is the exclusive use of bias-only tuning (BitFit~\cite{ZakenGR22}): every weight matrix in the transformer encoder remains frozen, and gradient updates are restricted to the bias vectors and the task-specific classification head.
This strategy yields extreme parameter efficiency, because \SystemNameO must store and transmit only the bias parameter vectors.
For a 12-layer Transformer, the aggregate number of trainable biases is below 0.08\% of the full parameter count~\cite{ZakenGR22}.
In every equation above (Eq.~(\ref{eq5}) through Eq.~(\ref{eq22})) the bias terms, highlighted in red, are additive and occupy a negligible fraction of the network capacity.

Relative to other parameter efficient fine tuning strategies such as LoRA~\cite{HuSWALWWC22} and adapter\cite{HoulsbyGJMLGAG19},\cite{HeZMBN22} approaches, BitFit constitutes the minimal invasive alternative because it restricts gradient updates to the bias vectors exclusively. This parsimony proves especially advantageous in small to medium scale data regimes where the objective is to adapt a pre trained language model to downstream tasks such as text classification or question answering~\cite{abs-2303-15647} without storing or optimising additional adapter matrices or low rank decompositions that would otherwise enlarge the inference model.

\subsubsection{\textbf{Alternating Attention}}
Drawing on recent analyses of efficient long-context modelling~\cite{abs-2408-00118,WarnerCCWHTGBLA25}, \SystemNameO alternates between two attention regimes  as illustrated in the RIGHT subfigure of Figure~\Ref{framework}: global attention, in which every token attends to every other token, and local sliding-window attention, in which interactions are confined to a fixed-width neighborhood, echoing the MBERT design.

Specifically, in the teacher model, global attention is applied every third layer with a RoPE (Rotary Position Embedding) theta value of 160,000, whereas the remaining layers use a local sliding window attention with 128 tokens and a RoPE theta value of 10,000. 
The student model replicates this schedule without alteration, inheriting the same theta assignments.
Consequently, the student's only departure from the teacher is the selective reuse of distinct layer subsets.
By inserting a global layer once every three Transformer blocks and constraining the remaining layers to a 128-token sliding window, \SystemNameO realises a substantial acceleration.

\subsubsection{\textbf{Flash Attention}}
Following the advancements detailed in~\cite{abs-2408-00118,WarnerCCWHTGBLA25}, our model incorporates Flash Attention, to enhance the efficiency of the Transformer. 
Although the initial Flash Attention 3 version lacked support for sliding window mechanisms, we adopt a similar strategy to MBERT by utilizing Flash Attention 3~\cite{ShahBZTRD24} (designed for H100 GPUs) for global attention layers and Flash Attention 2~\cite{Dao24} for local attention layers, as shown in Figure~\ref{framework}. 
Flash Attention is an ensemble of algorithms and GPU kernels that fuse the attention's matrix products and softmax into a tiled, on-chip pipeline, yielding exact but markedly faster and more memory-efficient execution than standard dense attention.

\subsubsection{\textbf{Sequence Unpadding}} 
\SystemNameO aligns with MosaicBERT~\cite{PortesTHKVNSKF23}, mGTE~\cite{ZhangZLXDTLYXHZ24}, and MBERT~\cite{WarnerCCWHTGBLA25} by implementing sequence unpadding in both training and inference stages.
Traditionally, encoder-only language models use padding tokens to standardize sequence lengths within a batch, leading to wasted computations on semantically meaningless tokens. 
Sequence unpadding addresses this by removing these padding tokens, combining sequences from a minibatch into a singular sequence, and processing this sequence as if it were a single batch. 
This method is particularly notable for its full-process integration, which eliminates padding tokens in both training and inference, thus avoiding unnecessary calculations. 

\subsubsection{\textbf{Embedding Frozen}}
`Freezing embeddings' keeps the embedding layer's parameters unchanged during backpropagation. While fine-tuning, most or all of a pre-trained model's parameters are adjusted for task-specific needs, but freezing embeddings means only the model's non-embedding parameters are updated. This technique assumes that the pre-trained embeddings have already captured universal data features applicable to multiple tasks.
Given the risk of overfitting inherent in fine-tuning large-scale pre-trained models with limited supervised data, we adopt a strategy of freezing the embedding layer. This approach is particularly advantageous in parameter-efficient training regimes. For instance, when fine-tuning BERT-base, freezing the embedding layer reduces the computational cost of gradient calculation by approximately 15\% per training batch.
Furthermore, this strategy introduces an implicit L2 regularization effect on the embedding parameters, thereby enhancing generalization and mitigating overfitting in downstream tasks with small datasets.

\begin{table*}[t]
  \centering
  \footnotesize
    \caption{Test accuracy (Acc (\%)) for semi-supervised text classification tasks and the baseline results are derived from \SystemNameA and \SystemNameB used in LiteSSLHub.} 
  \renewcommand\arraystretch{1.3}
  \setlength{\tabcolsep}{3.9mm}{
    \begin{tabular}{|l|ccc|ccc|ccc|c|}
   \specialrule{1pt}{0pt}{0pt}
  \multirow{2}{*}{\textbf{Models}} 
  & \multicolumn{3}{c|}{\textbf{AG News}}                                          
  & \multicolumn{3}{c|}{\textbf{Yahoo!Answer}}                                  
  & \multicolumn{3}{c|}{\textbf{DBpedia}}
  & \multicolumn{1}{c|}{\multirow{2}{*}{Avg}}                                   
  \\ \cline{2-10}
  & \multicolumn{1}{c|}{10}    & \multicolumn{1}{c|}{30}    & 200
  & \multicolumn{1}{c|}{10}    & \multicolumn{1}{c|}{30}    & 200   
  & \multicolumn{1}{c|}{10}    & \multicolumn{1}{c|}{30}    & 200
  &\multicolumn{1}{c|}{}  \\
  \hline
  MBERT\(_{\rm LARGE}\)
  & \multicolumn{1}{c|}{74.58}  & \multicolumn{1}{c|}{80.91}
  & \multicolumn{1}{c|}{88.75}  & \multicolumn{1}{c|}{52.18}  & \multicolumn{1}{c|}{62.42}  & \multicolumn{1}{c|}{70.42}  & \multicolumn{1}{c|}{91.52} & \multicolumn{1}{c|}{98.20} & 98.77 & \cellcolor[rgb]{0.851, 0.882, 0.957} 74.58 \\ 
   MBERT\(_{\rm LARGE}\) w. BitFit
   & \multicolumn{1}{c|}{78.97}  & \multicolumn{1}{c|}{84.62}  & 88.95
  & \multicolumn{1}{c|}{56.06}  & \multicolumn{1}{c|}{63.97}  & 69.11  & \multicolumn{1}{c|}{92.47}  & \multicolumn{1}{c|}{97.36} & 98.76 & \cellcolor[rgb]{0.624, 0.710, 0.890} 81.03 \\
  \rowcolor{gray!20}\textcolor{blue}{\SystemNameO} (S\(^{\rm MBERT-A13}\))
  & \multicolumn{1}{c|}{82.48}  & \multicolumn{1}{c|}{85.08}  & \multicolumn{1}{c|}{88.90}
  & \multicolumn{1}{c|}{60.91}  & \multicolumn{1}{c|}{65.33}  & \multicolumn{1}{c|}{69.26}
  & \multicolumn{1}{c|}{94.00}  & \multicolumn{1}{c|}{97.90} & 98.54 &  \cellcolor[rgb]{0.569, 0.667, 0.875} 82.50 
  \\
    \rowcolor{gray!20}\textcolor{blue}{\SystemNameO} (S\(^{\rm MBERT-B13}\))
  & \multicolumn{1}{c|}{78.71}  & \multicolumn{1}{c|}{70.58}  & \multicolumn{1}{c|}{83.85}
  & \multicolumn{1}{c|}{46.49}  & \multicolumn{1}{c|}{62.07}  & \multicolumn{1}{c|}{65.96}
  & \multicolumn{1}{c|}{90.15}  & \multicolumn{1}{c|}{83.38} & 91.80 &  \cellcolor[rgb]{0.851, 0.882, 0.957}74.64
  \\ 
  \hline
  BERT\(_{\rm BASE}\)
  & \multicolumn{1}{c|}{81.00}
  & \multicolumn{1}{c|}{84.32}
  & 87.24
  & \multicolumn{1}{c|}{60.10}
  & \multicolumn{1}{c|}{64.13}  & 69.28  & \multicolumn{1}{c|}{96.59}  & \multicolumn{1}{c|}{98.21}  & 98.79
  & \cellcolor[rgb]{0.482, 0.784, 0.565} 82.18
   \\
  UDA
  & \multicolumn{1}{c|}{84.70}  & \multicolumn{1}{c|}{86.89}  & 88.56
  & \multicolumn{1}{c|}{64.28}  & \multicolumn{1}{c|}{67.70}  & 69.71  & \multicolumn{1}{c|}{98.13}  & \multicolumn{1}{c|}{98.67} & 98.85 & \cellcolor[rgb]{0.388, 0.745, 0.482} 84.17 \\
  BERT\(_{\rm BASE}\) w. BitFit
 & \multicolumn{1}{c|}{75.26}  & \multicolumn{1}{c|}{83.93}  & 86.93
  & \multicolumn{1}{c|}{56.18}  & \multicolumn{1}{c|}{61.95}  & 68.92  & \multicolumn{1}{c|}{95.83}  & \multicolumn{1}{c|}{98.24} & 98.83 & \cellcolor[rgb]{0.553, 0.812, 0.624} 80.67\\
  \rowcolor{gray!20}\textcolor{blue}{\SystemNameO} (S\(^{\rm MBERT-A4}\))
  & \multicolumn{1}{c|}{80.52}  & \multicolumn{1}{c|}{83.76}  & \multicolumn{1}{c|}{87.58}
  & \multicolumn{1}{c|}{59.79}  & \multicolumn{1}{c|}{64.73}  & \multicolumn{1}{c|}{68.44}
  & \multicolumn{1}{c|}{94.25}  & \multicolumn{1}{c|}{97.11} & 97.98 &  \cellcolor[rgb]{0.510, 0.796, 0.588} 81.57
  \\
   \rowcolor{gray!20}\textcolor{blue}{\SystemNameO} (S\(^{\rm MBERT-B4}\))
  & \multicolumn{1}{c|}{74.01}  & \multicolumn{1}{c|}{78.23}  & \multicolumn{1}{c|}{80.42}
  & \multicolumn{1}{c|}{46.10}  & \multicolumn{1}{c|}{52.40}  & \multicolumn{1}{c|}{52.32}
  & \multicolumn{1}{c|}{83.35}  & \multicolumn{1}{c|}{87.84} & 86.73 &  \cellcolor[rgb]{0.988, 0.988, 1.000} 71.27 \\
  \hline
  TinyBERT\(^{6}\)        
  & \multicolumn{1}{c|}{71.45}
  & \multicolumn{1}{c|}{82.46}
  & 87.59
  & \multicolumn{1}{r|}{52.84}
  & \multicolumn{1}{c|}{60.59}
  & 68.71
  & \multicolumn{1}{c|}{96.89}
  & \multicolumn{1}{c|}{98.16}
  & 98.65
  & \cellcolor[rgb]{0.698, 0.769, 0.914}  79.70
  \\
  UDA\(\rm_{TinyBERT^{6}}\)
  & \multicolumn{1}{c|}{73.90}
  & \multicolumn{1}{c|}{85.16}
  & 87.54
  & \multicolumn{1}{c|}{57.14}
  & \multicolumn{1}{c|}{62.86}
  & 67.93
  & \multicolumn{1}{c|}{97.41}
  & \multicolumn{1}{c|}{97.87}
  & 98.26
  & \cellcolor[rgb]{0.412, 0.553, 0.835}   81.79
  \\
  \SystemNameA (S\(^{\rm A6}\))     
  & \multicolumn{1}{c|}{77.45}
  & \multicolumn{1}{c|}{86.93}
  & 88.82
  & \multicolumn{1}{c|}{59.10}
  & \multicolumn{1}{c|}{66.58}
  & 69.75
  & \multicolumn{1}{c|}{98.57}
  & \multicolumn{1}{c|}{98.61}
  & 98.73
  & \cellcolor[rgb]{0.282, 0.455, 0.796}  82.73
  \\
  \SystemNameA (S\(^{\rm B6}\))      
  & \multicolumn{1}{c|}{74.38}
  & \multicolumn{1}{c|}{86.39}
  & 88.70
  & \multicolumn{1}{c|}{57.62}
  & \multicolumn{1}{c|}{64.04}
  & 69.57
  & \multicolumn{1}{c|}{98.50}
  & \multicolumn{1}{c|}{98.45}
  & 98.57
  & \cellcolor[rgb]{0.380, 0.529, 0.824}  82.02
  \\
  \rowcolor{gray!20}\textcolor{blue}{\SystemNameO} (S\(^{\rm BERT-A6}\))
  & \multicolumn{1}{c|}{81.76}  & \multicolumn{1}{c|}{86.04}  & \multicolumn{1}{c|}{88.01}
  & \multicolumn{1}{c|}{62.65}  & \multicolumn{1}{c|}{65.90}  & \multicolumn{1}{c|}{68.51}
  & \multicolumn{1}{c|}{97.51}  & \multicolumn{1}{c|}{98.33}  & \multicolumn{1}{c|}{98.62} & \cellcolor[rgb]{0.541, 0.651, 0.871}   80.85 \\
   \rowcolor{gray!20}\textcolor{blue}{\SystemNameO} (S\(^{\rm BERT-B6}\))
   & \multicolumn{1}{c|}{81.71}  & \multicolumn{1}{c|}{84.40}  & \multicolumn{1}{c|}{86.43}
  & \multicolumn{1}{c|}{58.70}  & \multicolumn{1}{c|}{61.28}  & \multicolumn{1}{c|}{64.41}
  & \multicolumn{1}{c|}{94.91}  & \multicolumn{1}{c|}{96.01}  & \multicolumn{1}{c|}{96.92} & \cellcolor[rgb]{0.851, 0.882, 0.957}  78.56 \\
   \hline
  TinyBERT\(^{4}\)                         & \multicolumn{1}{c|}{69.67}
  & \multicolumn{1}{c|}{78.35}
  & 85.12
  & \multicolumn{1}{c|}{42.66}
  & \multicolumn{1}{c|}{53.63}
  & 61.89
  & \multicolumn{1}{c|}{89.65}
  & \multicolumn{1}{c|}{96.88}
  & 97.58
  & \cellcolor[rgb]{0.945, 0.973, 0.961} 75.05
  \\
  UDA\(\rm_{TinyBERT^{4}}\)
  & \multicolumn{1}{c|}{69.60}
  & \multicolumn{1}{c|}{77.56}
  & 83.60
  & \multicolumn{1}{c|}{40.69}
  & \multicolumn{1}{c|}{55.43}
  & 63.34
  & \multicolumn{1}{c|}{88.50}
  & \multicolumn{1}{c|}{93.63}
  & 95.98
  & \cellcolor[rgb]{0.988, 0.988, 1.000} 74.26 
  \\
  \SystemNameA (S\(^{\rm A4}\))
  & \multicolumn{1}{c|}{77.36}
  & \multicolumn{1}{c|}{85.55}
  & 87.95
  & \multicolumn{1}{c|}{51.31}
  & \multicolumn{1}{c|}{62.93}
  &  68.24
  & \multicolumn{1}{c|}{94.79}
  & \multicolumn{1}{c|}{98.14}
  & 98.63
  &  \cellcolor[rgb]{0.620, 0.839, 0.682}  80.54 
  \\
  \SystemNameA (S\(^{\rm B4}\))                
  & \multicolumn{1}{c|}{76.90}
  & \multicolumn{1}{c|}{85.39}
  &  87.82
  & \multicolumn{1}{c|}{51.48}
  & \multicolumn{1}{c|}{62.36}
  &  68.10
  & \multicolumn{1}{c|}{94.02}
  & \multicolumn{1}{c|}{98.13}
  &  98.56
  & \cellcolor[rgb]{0.631, 0.843, 0.694}  80.31
  \\
   \SystemNameB (S\(^{\rm A4}\)) 
  & \multicolumn{1}{c|}{81.03}
  & \multicolumn{1}{c|}{87.32}
  & 89.04
  & \multicolumn{1}{c|}{62.33}
  & \multicolumn{1}{c|}{68.10}
  & 71.35
  & \multicolumn{1}{c|}{97.19}
  & \multicolumn{1}{c|}{98.70}
  & 98.90
  & \cellcolor[rgb]{0.427, 0.761, 0.518} 83.77 \\
  \SystemNameB (S\(^{\rm B4}\))
  & \multicolumn{1}{c|}{82.06}
  & \multicolumn{1}{c|}{87.38}
  & 89.77
  & \multicolumn{1}{c|}{65.21}
  & \multicolumn{1}{c|}{68.02}
  &  71.08
  & \multicolumn{1}{c|}{98.44}
  & \multicolumn{1}{c|}{98.71}
  & 98.82
  & \cellcolor[rgb]{0.388, 0.745, 0.482}  84.39  \\
  \rowcolor{gray!20}\textcolor{blue}{\SystemNameO} (S\(^{\rm BERT-A4}\))
  & \multicolumn{1}{c|}{81.67}  & \multicolumn{1}{c|}{85.83}  & \multicolumn{1}{c|}{87.80}
  & \multicolumn{1}{c|}{58.22}  & \multicolumn{1}{c|}{66.19}  & \multicolumn{1}{c|}{68.30}
  & \multicolumn{1}{c|}{96.48}  & \multicolumn{1}{c|}{97.50}  & \multicolumn{1}{c|}{98.16} & \cellcolor[rgb]{0.694, 0.871, 0.745}  79.26 \\
   \rowcolor{gray!20}\textcolor{blue}{\SystemNameO} (S\(^{\rm BERT-B4}\))
  & \multicolumn{1}{c|}{78.56}  & \multicolumn{1}{c|}{83.70}  & \multicolumn{1}{c|}{85.23}
  & \multicolumn{1}{c|}{62.99}  & \multicolumn{1}{c|}{61.73}  & \multicolumn{1}{c|}{64.46}
  & \multicolumn{1}{c|}{95.28}  & \multicolumn{1}{c|}{96.57}  & \multicolumn{1}{c|}{96.99} &  \cellcolor[rgb]{0.718, 0.878, 0.765}  78.87 \\
  \hline
  FLiText                              & \multicolumn{1}{c|}{67.14}
  & \multicolumn{1}{c|}{77.12}
  & 82.12
  & \multicolumn{1}{c|}{48.30}
  & \multicolumn{1}{c|}{57.01}
  & 63.09
  & \multicolumn{1}{c|}{89.26}
  & \multicolumn{1}{c|}{94.04}
  & 97.01
  & \cellcolor[rgb]{0.851, 0.882, 0.957}  75.01 \\
    \SystemNameA (S\(^{\rm A2}\))  
  & \multicolumn{1}{c|}{75.05}
  & \multicolumn{1}{c|}{82.16}
  & 86.38
  & \multicolumn{1}{c|}{51.05}
  & \multicolumn{1}{c|}{58.83}
  & 65.63
  & \multicolumn{1}{c|}{89.55}
  & \multicolumn{1}{c|}{96.14}
  & 97.70
  & \cellcolor[rgb]{0.659, 0.737, 0.902} 78.05 \\
  \SystemNameA (S\(^{\rm B2}\))           
  & \multicolumn{1}{c|}{70.61}
  & \multicolumn{1}{c|}{81.87}
  &  86.08
  & \multicolumn{1}{c|}{48.41}
  & \multicolumn{1}{c|}{57.84}
  &  64.04
  & \multicolumn{1}{c|}{89.67}
  & \multicolumn{1}{c|}{96.06}
  &  97.58
  & \cellcolor[rgb]{0.729, 0.792, 0.925}  76.90 
  \\ 
  \SystemNameB (S\(^{\rm A2}\)) 
  & \multicolumn{1}{c|}{81.14}
  & \multicolumn{1}{c|}{85.35}
  & 87.10
  & \multicolumn{1}{c|}{61.12}
  & \multicolumn{1}{c|}{64.40}
  &  66.33
  & \multicolumn{1}{c|}{96.61}
  & \multicolumn{1}{c|}{98.24}
  & 98.33
  & \cellcolor[rgb]{0.400, 0.541, 0.831}  82.07  \\
  \SystemNameB (S\(^{\rm B2}\)) 
  & \multicolumn{1}{c|}{81.89}
  & \multicolumn{1}{c|}{87.69}
  & 89.11
  & \multicolumn{1}{c|}{64.16}
  & \multicolumn{1}{c|}{66.88}
  &  69.57
  & \multicolumn{1}{c|}{98.05}
  & \multicolumn{1}{c|}{98.77}
  & 98.57
  & \cellcolor[rgb]{0.282, 0.455, 0.796}  83.85  \\
  \rowcolor{gray!20}\textcolor{blue}{\SystemNameO} (S\(^{\rm BERT-A2}\))
  & \multicolumn{1}{c|}{83.47}  & \multicolumn{1}{c|}{85.32}  & \multicolumn{1}{c|}{86.67}
  & \multicolumn{1}{c|}{62.47}  & \multicolumn{1}{c|}{66.11}  & \multicolumn{1}{c|}{64.44}
  & \multicolumn{1}{c|}{93.10}  & \multicolumn{1}{c|}{95.28}  & \multicolumn{1}{c|}{94.42} & \cellcolor[rgb]{0.553, 0.659, 0.875}  79.67  \\
  \rowcolor{gray!20}\textcolor{blue}{\SystemNameO} (S\(^{\rm BERT-B2}\))
    & \multicolumn{1}{c|}{79.58}  & \multicolumn{1}{c|}{85.24}  & \multicolumn{1}{c|}{83.44}
  & \multicolumn{1}{c|}{58.21}  & \multicolumn{1}{c|}{61.47}  & \multicolumn{1}{c|}{68.41}
  & \multicolumn{1}{c|}{93.52}  & \multicolumn{1}{c|}{94.52}  & \multicolumn{1}{c|}{95.89} & \cellcolor[rgb]{0.612, 0.702, 0.890}  78.77 \\
  \specialrule{1pt}{0pt}{0pt}
  \end{tabular}
  }
  \label{main_text_classification}
\end{table*}

\section{Experiments}
\subsection{Implementation Detail}
\subsubsection{Hyperparameters of \SystemNameO with BERT Backbone}
By adopting BERT$_{\text{BASE}}$ as the teacher backbone, we distill a lightweight student that retains merely two of the twelve transformer layers, achieving a parameter budget comparable to \SystemNameA and \SystemNameB.
BERT\(_{BASE}\) as the teacher model, encapsulates 109 million parametersconfigured by \(N=12\)  layers, hidden dimension \(d=768\), feed-forward dimension \(d=3072\), and \(h=12\) attention heads. 
Text tokenization is conducted using the BERT tokenizer, with maximum sentence lengths set to   256 for text classification. 
The student BERT skeleton retains the dropout schema prescribed by prior distillation studies: token dropout probability 0.2 and hidden dropout probability 0.1, values cross-validated in Yan et al.~\cite{YanLWZWX20} and Jiang et al.~\cite{JiangMLLDYW23}.
Optimisation proceeds via Adam with first- and second-order momentum coefficients 0.9 and 0.999, respectively.
A single learning-rate schedule—initial value 5e-4 with linear decay—is shared by the supervised cross-entropy, unsupervised mean-squared-error, and mutual-learning objectives; the first 20\% of iterations are reserved for linear warm-up.
A balance factor for supervised and unsupervised learning is consistently set at 1. 
Mini-batch construction assigns 4 to the labeled stream and 16 to the unlabeled stream in all classification scenarios.
Consistent with \SystemNameB, the architectural depth of the student is frozen at the configuration previously identified as optimal, thereby eliminating depth-driven variance from the experimental design:

\(\clubsuit\) \textbf{6-layer students S\(^{\rm BERT-x6}\):}
\begin{itemize}
  \item Series A6: the first 6 layers, indices [0-5].
  \item Series B6: the last 6 layers, indices [6-11].
\end{itemize}

\(\clubsuit\) \textbf{4-layer students S\(^{\rm BERT-x4}\):}
\begin{itemize}
  \item Series A4: the first 4 layers, indices [0-3].
  \item Series B4: the last 4 layers, indices [8-11].
\end{itemize}

\(\clubsuit\) \textbf{2-layer students S\(^{\rm BERT-x2}\):}
\begin{itemize}
  \item Series A2: the first 2 layers, indices [0-1].
  \item Series B2: the last 2 layers, indices [10-11].
\end{itemize}

\subsubsection{Hyperparameters of \SystemNameO with MBERT Backbone}
ModernBERT eliminates biases in all Layer Norm and Linear layers except the task-specific head to improve inference efficiency.  
In our system, we revert this design by re-introducing biases to every Linear and Layer Normalization module.   
During fine-tuning, the embedding matrix is frozen to reduce representation drift, thereby enhancing training stability and convergence speed, especially in low-resource adaptation scenarios.
The model is optimized with Decoupled AdamW~\cite{LoshchilovH19} (lr=$1.0{\times}10^{-3}$, $\beta_1\!=\!0.9$, $\beta_2\!=\!0.98$, $\varepsilon\!=\!1.0{\times}10^{-6}$, weight-decay=$1.0{\times}10^{-6}$) to separate $\ell_2$ regularization from gradient updates, yielding improved generalization stability. A linear-decay schedule~\cite{LoshchilovH19} with Warmup-Stable-Decay (WSD)~\cite{Zhai0HB22,abs-2404-06395,WarnerCCWHTGBLA25} is employed: $6\%$ of the total steps are allocated to linear warm-up to the peak learning rate, followed by linear decay to $2\%$ of the maximum. Training proceeds for 3 epochs with batch size 32 in bfloat16~\cite{AnthonyHNBBYSS024} mixed precision, preserving numerical stability while reducing memory and computational overhead. The maximum sequence length is 512 and longer sequences are truncated consistent with BERT pre-training. All experiments are conducted on a single NVIDIA L20 48GB GPU.  MBERT\(_{LARGE}\) is deep-and-narrow (28 layers) with hidden sizes (1024) to multiples of 64 for efficient tensor-core utilization on standard GPUs.

\subsubsection{Student-Model Construction Strategy}
To align the initial architecture of the student model with the pre-trained layer-wise pattern of the teacher model—especially the alternating global-attention distribution and RoPE configuration—we adopt the following layer-selection policy. 
ModernBERT-Large contains 28 layers: layer 0 uses global attention, and every second layer from 1 to 27 also employs global attention. RoPE $\theta$: identical to ModernBERT; $\theta = 160,000$ for global-attention layers and $\theta = 10,000$ for all others.

\(\spadesuit\)  \textbf{13-layer students S\(^{\rm MBERT-x13}\):}
\begin{itemize}
  \item A13: the first 13 layers, indices [0-12].
  \item B13: layer 0 \& the last 12 layers, indices [0, 16-27].
\end{itemize}

\(\spadesuit\) \textbf{4-layer students S\(^{\rm MBERT-x4}\):}
\begin{itemize}
  \item A4: the first 4 layers, indices [0-3].
  \item B4: layer 0 \& the last 3 layers, indices [0, 25-27].
\end{itemize}


\subsection{Datasets}


We evaluate on five semi-supervised text classification benchmarks adopted by \SystemNameA and \SystemNameB: AG News~\cite{ZhangZL15}, Yahoo! Answers~\cite{ChangRRS08}, and DBpedia~\cite{mendes2012dbpedia}.
Models are trained with 10, 30, and 200 labeled examples per class as reported in Table~\ref{statistics}.
These corpora are obtained from LiteSSLHub\footnote{\url{https://github.com/LiteSSLHub}}.
To extend the evaluation to sentiment-oriented SSL we additionally employ the USB collection which supplies Amazon~\cite{McAuleyL13} and Yelp\footnote{\url{https://www.yelp.com/dataset}} reviews at 50 and 200 labels per class.
The USB release also re-packages AG News and Yahoo! Answers that were previously used in~\cite{ChenYY20,XieDHL020,McAuleyL13} and these corpora remain challenging for SSL.
Amazon and Yelp reviews provide fine-grained sentiment labels for stress-testing SSL algorithms.
These datasets are obtained from USB\footnote{\url{https://github.com/microsoft/Semi-supervised-learning }}. 
As illustrated in Table~\ref{statistics}, the identical datasets originate from different repositories and differ only in the samples of labels and in the sizes of the validation and test sets, thus there is no essential distinction.

\begin{table*}[t]
  \centering
  \footnotesize
  \caption{Test accuracy (Acc (\%)) for semi-supervised text classification tasks and the baseline results are derived from USB benchmark and its github  repository (\(*\)).  Each setting runs 3 different random seeds and computes the average performance with standard deviation.}
   \renewcommand\arraystretch{1.5}
  \setlength{\tabcolsep}{1.3mm}{
  \begin{tabular}{|l|cc|cc|cc|cc|cc|c|}
    \specialrule{1pt}{0pt}{0pt}
    \multirow{2}{*}{\textbf{Models}} 
    & \multicolumn{2}{c|}{\textbf{Agnews}}                                          
    & \multicolumn{2}{c|}{\textbf{Amazon}}                                  
    & \multicolumn{2}{c|}{\textbf{Yahoo}}
    & \multicolumn{2}{c|}{\textbf{Yelp}}
   &\multicolumn{1}{c|}{\multirow{2}{*}{Avg}} \\ \cline{2-9}
    & \multicolumn{1}{c|}{40}    & \multicolumn{1}{c|}{200}    
    & \multicolumn{1}{c|}{250}   & \multicolumn{1}{c|}{1000}    
    & \multicolumn{1}{c|}{500}   & \multicolumn{1}{c|}{2000}    
    & \multicolumn{1}{c|}{250}   & \multicolumn{1}{c|}{1000} 
    &\multicolumn{1}{c|}{}  \\
    \hline
Fully-Supervised  & 94.26$\pm$0.30 & 94.36$\pm$0.05 & 63.19$\pm$0.05 & 63.12$\pm$0.19 & 73.75$\pm$1.07 & 74.45$\pm$0.43 & 68.26$\pm$0.23 & 67.30$\pm$0.58 &   \multicolumn{1}{c|}{\cellcolor[rgb]{0.388, 0.745, 0.482} 79.84} \\
Fully-Supervised*  & 94.22$\pm$0.07 & 94.27$\pm$0.11 & 63.60$\pm$0.05 & 63.60$\pm$0.05 & 75.13$\pm$0.04 & 75.16$\pm$0.04 & 67.96$\pm$0.03 & 67.96$\pm$0.03 &  \multicolumn{1}{c|}{\cellcolor[rgb]{0.392, 0.749, 0.486} 79.61} \\
Supervised  & 84.99$\pm$1.21 & 87.00$\pm$1.00 & 48.26$\pm$0.63 & 52.66$\pm$0.66 & 62.90$\pm$1.22 & 66.44$\pm$0.08 & 49.73$\pm$0.51 & 53.04$\pm$0.42 &      \multicolumn{1}{c|}{\cellcolor[rgb]{0.541, 0.808, 0.616} 70.63} \\
Supervised*   & 84.94$\pm$1.08 & 85.75$\pm$0.97 & 47.69$\pm$1.28 & 52.47$\pm$0.69 & 62.57$\pm$0.29 & 66.74$\pm$0.10 & 48.78$\pm$0.98 & 53.29$\pm$0.37 &  \multicolumn{1}{c|}{\cellcolor[rgb]{0.549, 0.812, 0.620} 70.28} \\
P-Model   & 53.16$\pm$6.20 & 86.54$\pm$0.76 & 26.47$\pm$6.92 & 51.73$\pm$0.48 & 58.63$\pm$2.15 & 67.04$\pm$0.16 & 26.65$\pm$2.31 & 47.98$\pm$1.48 &   \multicolumn{1}{c|}{\cellcolor[rgb]{0.729, 0.886, 0.776} 59.28} \\
Pseudo-label   & 76.14$\pm$7.63 & 87.71$\pm$0.40 & 47.00$\pm$1.48 & 53.51$\pm$0.45 & 61.40$\pm$1.09 & 66.56$\pm$0.24 & 44.30$\pm$0.95 & 52.28$\pm$0.37 &  \multicolumn{1}{c|}{\cellcolor[rgb]{0.576, 0.824, 0.643} 68.61} \\
Pseudo-label*   & 81.51$\pm$3.07 & 85.31$\pm$1.88 & 46.55$\pm$1.90 & 53.00$\pm$0.79 & 62.30$\pm$0.65 & 67.28$\pm$0.31 & 45.49$\pm$0.82 & 52.67$\pm$0.20 &    \multicolumn{1}{c|}{\cellcolor[rgb]{0.565, 0.820, 0.635} 69.26} \\
MeanTeacher   & 85.02$\pm$1.10 & 86.77$\pm$1.12 & 48.33$\pm$0.45 & 52.49$\pm$0.24 & 63.03$\pm$1.02 & 66.57$\pm$0.22 & 48.93$\pm$1.44 & 53.39$\pm$0.34 & 
 \multicolumn{1}{c|}{\cellcolor[rgb]{0.545, 0.808, 0.616} 70.57} \\
MeanTeacher*   & 84.83$\pm$1.21 & 86.07$\pm$0.52 & 47.86$\pm$0.52 & 52.34$\pm$0.84 & 62.91$\pm$0.18 & 66.47$\pm$0.28 & 49.40$\pm$0.62 & 52.79$\pm$0.31 &  \multicolumn{1}{c|}{\cellcolor[rgb]{0.553, 0.812, 0.624} 70.08} \\
VAT   & 85.00$\pm$1.12 & 88.41$\pm$0.94 & 49.62$\pm$0.83 & 53.96$\pm$0.28 & 64.84$\pm$0.41 & 68.47$\pm$0.41 & 47.24$\pm$0.87 & 54.47$\pm$0.13 &  \multicolumn{1}{c|}{\cellcolor[rgb]{0.537, 0.808, 0.612} 70.88} \\
VAT*   & 85.30$\pm$1.19 & 88.29$\pm$0.84 & 50.17$\pm$0.46 & 53.46$\pm$0.31 & 65.13$\pm$0.41 & 68.50$\pm$0.55 & 47.03$\pm$1.41 & 54.70$\pm$0.32 &  \multicolumn{1}{c|}{\cellcolor[rgb]{0.549, 0.812, 0.620} 70.32} \\
UDA   & 81.27$\pm$2.68 & 87.66$\pm$1.90 & 47.52$\pm$1.20 & 54.49$\pm$0.61 & 64.69$\pm$0.43 & 67.99$\pm$0.68 & 41.78$\pm$0.40 & 57.82$\pm$0.68 &  \multicolumn{1}{c|}{\cellcolor[rgb]{0.545, 0.812, 0.620} 70.40} \\
Fixmatch   & 77.20$\pm$5.18 & 88.57$\pm$0.65 & 52.15$\pm$1.22 & 56.27$\pm$0.45 & 65.85$\pm$0.94 & 69.24$\pm$0.53 & 49.66$\pm$0.40 & 58.01$\pm$0.58 &  \multicolumn{1}{c|}{\cellcolor[rgb]{0.580, 0.824, 0.647} 68.41} \\
Fixmatch*   & 69.83$\pm$1.87 & 88.29$\pm$1.95 & 52.39$\pm$0.83 & 56.95$\pm$0.54 & 66.97$\pm$0.49 & 69.49$\pm$0.53 & 53.48$\pm$0.94 & 59.35$\pm$0.46 &  \multicolumn{1}{c|}{\cellcolor[rgb]{0.561, 0.816, 0.631} 69.41} \\
Flexmatch   & 83.10$\pm$6.76 & 88.57$\pm$0.91 & 54.25$\pm$1.21 & 56.86$\pm$0.82 & 64.19$\pm$1.09 & 68.58$\pm$0.41 & 53.63$\pm$0.74 & 59.14$\pm$0.74 & 
     \multicolumn{1}{c|}{\cellcolor[rgb]{0.612, 0.835, 0.675} 66.45} \\
Flexmatch*   & 83.62$\pm$3.94 & 87.92$\pm$0.73 & 54.27$\pm$1.60 & 57.75$\pm$0.33 & 64.39$\pm$1.08 & 68.87$\pm$0.18 & 56.65$\pm$0.69 & 59.49$\pm$0.34 &  \multicolumn{1}{c|}{\cellcolor[rgb]{0.596, 0.831, 0.659} 67.46} \\
DASH*      & $68.33\pm3.19$ & $86.24\pm1.67$ & $52.90\pm0.74$ & $56.91\pm0.60$ & $64.74\pm0.33$ & $68.81\pm0.29$ & $54.76\pm2.02$ & $59.86\pm0.79$ & \multicolumn{1}{c|}{\cellcolor[rgb]{0.643, 0.851, 0.702} 64.53} \\ 
Crmatch*    & $87.72\pm1.43$ & $88.92\pm1.24$ & $54.51\pm0.98$ & $56.93\pm0.50$ & $67.49\pm0.40$  & $70.02\pm0.07$ & $54.29\pm0.63$ & $59.38\pm0.28$  & \multicolumn{1}{c|}{\cellcolor[rgb]{0.576, 0.824, 0.643} 68.65}\\
AdaMatch   & 87.08$\pm$1.53 & 88.97$\pm$0.62 & 53.25$\pm$1.23 & 56.50$\pm$0.67 & 67.03$\pm$0.43 & 69.18$\pm$0.29 & 51.84$\pm$0.80 & 58.29$\pm$1.00 &  \multicolumn{1}{c|}{\cellcolor[rgb]{0.580, 0.824, 0.647} 68.35} \\
AdaMatch*   & 88.27$\pm$0.17 & 88.78$\pm$0.95 & 53.28$\pm$0.72 & 57.73$\pm$0.25 & 67.25$\pm$0.35 & 69.56$\pm$0.31 & 54.60$\pm$0.96 & 59.84$\pm$0.49 & \multicolumn{1}{c|}{\cellcolor[rgb]{0.573, 0.824, 0.643} 68.68} \\
SimMatch   & 85.20$\pm$0.57 & 88.88$\pm$0.15 & 52.73$\pm$1.73 & 56.91$\pm$0.50 & 65.85$\pm$0.9 & 69.36$\pm$0.42 & 53.60$\pm$1.71 & 58.76$\pm$0.17 & \multicolumn{1}{c|}{\cellcolor[rgb]{0.600, 0.831, 0.667} 67.06} \\
Simmatch*   & 85.74$\pm$1.51 & 87.55$\pm$1.37 & 54.09$\pm$0.95 & 57.79$\pm$0.30 & 66.94$\pm$0.2 & 69.84$\pm$0.21 & 53.88$\pm$0.48 & 59.74$\pm$0.62 & \multicolumn{1}{c|}{\cellcolor[rgb]{0.584, 0.827, 0.651} 68.02} \\
Comatch*    & $88.05\pm0.76$ & $89.25\pm0.35$ & $51.24\pm0.90$ & $56.64\pm0.21$ & $66.52\pm0.51$ & $69.75\pm0.35$ & $54.60\pm1.12$ & $59.73\pm0.51$ &  \multicolumn{1}{c|}{\cellcolor[rgb]{0.604, 0.831, 0.667} 66.97} \\ 
Freematch*   & \(87.02\pm0.58\) & \(88.27\pm0.63\) & \(53.59\pm0.60\) & \(57.36\pm0.06\) & \(67.23\pm0.26\) & \(69.68\pm0.18\) & \(52.05\pm1.45\) & \(59.63\pm1.0\) & \multicolumn{1}{c|}{\cellcolor[rgb]{0.612, 0.835, 0.675} 66.48} \\
Softmatch*   & \(88.10\pm0.27\) & \(88.28\pm1.58\) & \(54.71\pm0.95\) & \(57.79\pm0.20\) & \(66.93\pm0.31\) & \(69.56\pm0.62\) & \(55.91\pm0.5\) & \(60.24\pm0.13\) & \multicolumn{1}{c|}{\cellcolor[rgb]{0.592, 0.827, 0.659} 67.69} \\ 
\hline
\SystemNameA (S\(^{\rm A2}\))  & 75.33$\pm$0.72&  84.89$\pm$0.55&  30.41$\pm$0.99&  43.71$\pm$0.81&  55.74$\pm$0.77&  62.58$\pm$0.64&  31.98$\pm$1.15&  45.37$\pm$0.68&  \multicolumn{1}{c|}{\cellcolor[rgb]{0.839, 0.929, 0.871} 52.65} \\
\SystemNameA (S\(^{\rm B2}\))   & 72.57$\pm$0.61&  85.01$\pm$0.73&  29.09$\pm$1.02&  42.90$\pm$0.85&  54.70$\pm$0.90&  60.92$\pm$0.71&  31.18$\pm$0.83&  44.96$\pm$1.04&   \multicolumn{1}{c|}{\cellcolor[rgb]{0.855, 0.933, 0.886} 51.73} \\
\SystemNameB (S\(^{\rm A2}\))  & 76.12$\pm$0.68& 85.34$\pm$0.51& 31.05$\pm$0.94& 44.23$\pm$0.78& 56.18$\pm$0.74& 63.11$\pm$0.61& 32.47$\pm$1.09& 45.89$\pm$0.65& \multicolumn{1}{c|}{\cellcolor[rgb]{0.839,0.929,0.871} 53.70} \\
\SystemNameB (S\(^{\rm B2}\))  & 73.24$\pm$0.58& 85.42$\pm$0.69& 29.76$\pm$0.97& 43.51$\pm$0.81& 55.03$\pm$0.86& 61.35$\pm$0.67& 31.73$\pm$0.79& 45.38$\pm$0.98& \multicolumn{1}{c|}{\cellcolor[rgb]{0.855,0.933,0.886} 52.43} \\
\rowcolor{gray!20}\textcolor{blue}{\SystemNameO} (S\(^{\rm MBERT-A4}\))
 & 83.20$\pm$1.42 & 87.05$\pm$1.28 & 42.46$\pm$0.88 & 49.92$\pm$0.27 & 63.65$\pm$0.18 & 66.37$\pm$0.19 & 50.15$\pm$0.44 & 54.68$\pm$0.57 & \multicolumn{1}{c|}{\cellcolor[rgb]{0.659, 0.855, 0.718} 63.46} \\
\rowcolor{gray!20}\textcolor{blue}{\SystemNameO} (S\(^{\rm MBERT-B4}\))
 & 78.32$\pm$1.44 & 80.94$\pm$1.31 & 34.18$\pm$0.87 & 39.52$\pm$0.28 & 52.64$\pm$0.17 & 52.26$\pm$0.20 & 40.85$\pm$0.43 & 45.50$\pm$0.59 &  
\multicolumn{1}{c|}{\cellcolor[rgb]{0.824, 0.922, 0.859} 53.61} \\ 
\rowcolor{gray!20}\textcolor{blue}{\SystemNameO} (S\(^{\rm BERT-A2}\))
 & 82.71$\pm$1.40 & 86.13$\pm$1.26 & 37.66$\pm$0.85 & 40.47$\pm$0.29 & 62.30$\pm$0.19 & 63.91$\pm$0.20 & 41.70$\pm$0.42 & 44.27$\pm$0.55 &  \multicolumn{1}{c|}{\cellcolor[rgb]{0.749, 0.894, 0.792} 58.15} \\
\rowcolor{gray!20}\textcolor{blue}{\SystemNameO} (S\(^{\rm BERT-B2}\))
 & 80.64$\pm$1.43 & 84.12$\pm$1.29 & 35.13$\pm$0.84 & 37.99$\pm$0.30 & 57.52$\pm$0.18 & 59.65$\pm$0.21 & 37.86$\pm$0.41 & 41.87$\pm$0.57 &  \multicolumn{1}{c|}{\cellcolor[rgb]{0.788, 0.910, 0.827} 55.75} \\
  \specialrule{1pt}{0pt}{0pt}
  \hline
\end{tabular}
}
\label{tab:performance_comparison}
\end{table*}

\subsection{Baselines and Comparison}
For text classification, we compare with (\(i\)) \textit{supervised baselines with FULL model}, BERT\(\rm_{BASE}\), MBERT\(\rm_{LARGE}\) and Supervised BERT in USB. 
(\(ii\)) \textit{semi-supervised baselines with FULL model}, like UDA~\cite{XieDHL020}, and  FLiText~\cite{LiuZFHL21} and 17 SSL models in USB. (\(iii\)) \textit{supervised baselines with SMALL model}, like TinyBERT~\cite{JiaoYSJCL0L20}. (\(vi\)) \textit{supervised FULL model with parameter-efficient training}, like MBERT\(\rm_{LARGE}\) w. BitFit,  BERT\(\rm_{BASE}\) w. BitFit.  (\(vi\)) \textit{semi-supervised baselines with SMALL model}, like FLiText,  UDA\(\rm_{TinyBERT^{6}}\), \SystemNameA~\cite{JiangMLLDYW23} and \SystemNameB~\cite{MaoJLLLWLL25}.

\SystemNameB is the state-of-the-art faster and lightweight semi-supervised framework which employs a co-training technique to optimize multiple small student models, promoting knowledge sharing among students through diverse data and model views.  MBERT$_{\rm LARGE}$ possesses the largest backbone among all models; however, it should be noted that \SystemNameO re-uses this same backbone. Consequently, as illustrated in Figure~\ref{comparing}, MBERT$_{\rm LARG}$ is a parameter-efficient model: it employs alternating attention, sequence unpadding, Flash Attention, bFloat16 tuning, and frozen embeddings during training (in contrast to our approach, it does not use only-bias tuning and lacks SSL, TOT, and DML semi-supervised strategies), and applies alternating attention, sequence unpadding, Flash Attention, and bFloat16 inference at prediction time (it does not incorporate our lightweight inference module).
MBERT$_{\rm LARGE}$ w. BitFit denotes a parameter-efficient training regime that updates only bias parameters.

We position our approach against the current best-performing semi-supervised text-classification methods on the Unified SSL Benchmark (USB)~\cite{wang2022usb}.
The benchmark’s reference implementation subsumes the following family of algorithms: Pmodel, MeanTeacher~\cite{TarvainenV17}, VAT~\cite{MiyatoMKI19}, MixMatch~\cite{BerthelotCGPOR19}, ReMixMatch~\cite{AimarHXKF24}, UDA~\cite{XieDHL020}, FixMatch~\cite{SohnBCZZRCKL20}, Dash~\cite{XuSYQLSLJ21}, CoMatch~\cite{0001XH21}, CRMatch~\cite{FanKS21}, FlexMatch~\cite{ZhangWHWWOS21}, AdaMatch~\cite{BerthelotRSCK22}, SimMatch~\cite{ZhengYHWQX22}, FreeMatch, and SoftMatch.
The majority of these SSL algorithms were originally developed for computer vision tasks. Wang et al.~\cite{wang2022usb} transfer them to natural language processing by inserting a twelve-layer BERT encoder as the backbone.
Contemporary semi-supervised learning minimizes the supervised cross-entropy objective and simultaneously imposes consistency constraints on unlabeled instances. FixMatch, FlexMatch, FreeMatch, and SoftMatch progressively refine pseudo-label selection—FixMatch uses a fixed threshold, FlexMatch adapts per-class thresholds by hit rate, FreeMatch further tunes thresholds online with an early-training fairness term, and SoftMatch maintains both coverage and accuracy to fully utilize unlabeled data without trade-offs.


\begin{figure*}[tbp]
\centering

\begin{minipage}[b]{1.0\textwidth}
  \centering
  \includegraphics[width=\linewidth]{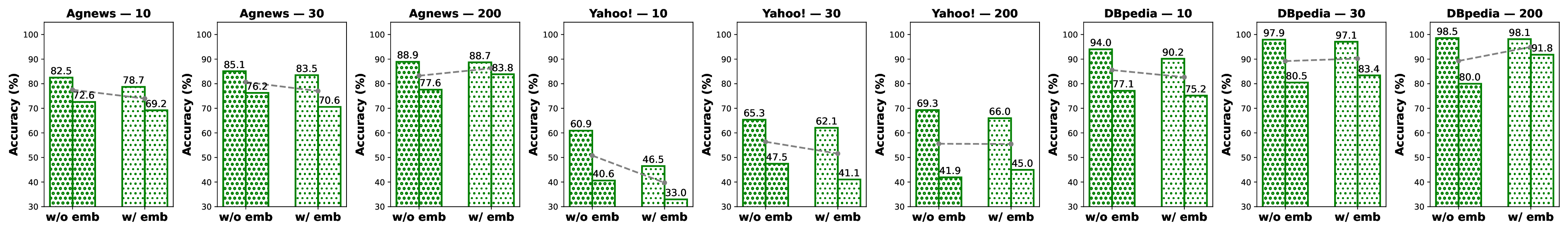}
  (a) {\scriptsize \SystemNameO (S\(^{\rm MBERT-x\textbf{13}}\)) with or without Tuning Embeddings}
\end{minipage}

\begin{minipage}[b]{1.0\textwidth}
  \centering
  \includegraphics[width=\linewidth]{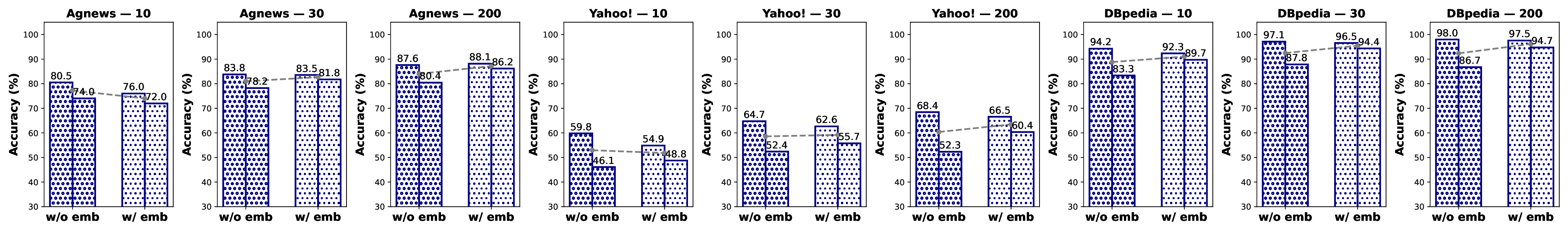}
  (b) {\scriptsize \SystemNameO (S\(^{\rm MBERT-x\textbf{4}}\)) with or without Tuning Embeddings}
\end{minipage}

\begin{minipage}[b]{1.0\textwidth}
  \centering
  \includegraphics[width=\linewidth]{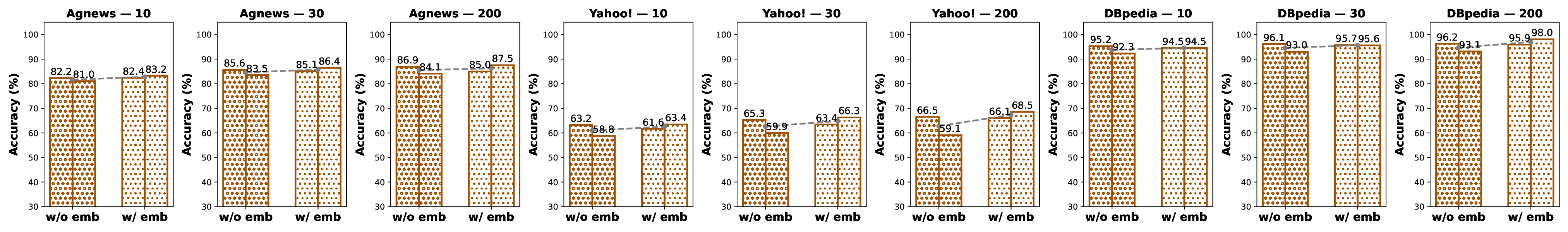}
  (c) {\scriptsize \SystemNameO (S\(^{\rm BERT-x\textbf{4}}\)) with or without Tuning Embeddings}
\end{minipage}

\caption{Freezing or unfreezing embeddings exerts a measurable influence on performance results. 
Adjacent bars represent the two student networks internal to \SystemNameO, and the figure superimposes the mean accuracy averaged across both students under frozen-embedding and unfrozen-embedding conditions. 
}
\label{framework_results}
\end{figure*}

\section{Experimental Results}
\subsection{Evaluation on Text Classification}
Table~\ref{main_text_classification} summarises the mean accuracy of  \SystemNameO variants. 
Within the 8.9M–9.6M parameter band, \SystemNameO (S\(^{\rm BERT-A2}\)) attains 84.47 on AgNews with 10 labels, ranking first and surpassing the second-best \SystemNameB by 1.58 percentage points. With only 101.71M parameters it yields results comparable to 396.83M and 396.18M MBERT\(_{\rm LARGE}\) with or without BitFit settings.
\SystemNameO (S\(^{\rm BERT-A2}\))  achieves performance comparable to or surpassing that of \SystemNameA and \SystemNameB while relying on parameter-efficient training.  Table~\ref{addlabel-parameter} confirms that fewer than 1M weights are updated, leaving the vast majority of the network frozen.

\SystemNameO (S\(^{\rm MBERT-B13}\)) retains the full 101.71 M parameter count yet registers an average accuracy 7.86\% points below that of \SystemNameO (S\(^{\rm MBERT-B4}\)).
An analogous disparity is observed between S\(^{\rm MBERT-A4}\) and S\(^{\rm MBERT-B4}\), where the former, despite identical parameter size, underperforms by 10.30\% points.
These pairwise comparisons demonstrate that representational quality is independent of teacher depth.  For the MBERT architecture, a shallow, lightly-parametrized  and effectively frozen encoder yields superior training signals relative to its deeper counterpart.

Table~\ref{main_text_classification} demonstrates that \SystemNameO (S\(^{\rm MBERT-A13}\)) surpasses its teacher MBERT\(_{\rm LARGE}\), attaining 82.50\% average accuracy and thereby exceeding both the bias-based lightweight MBERT\(_{\rm LARGE}\) and the standard full-parameter model.
This outcome demonstrates the architecture's superior capacity for semi-supervised learning under scarce labels. 
As the student depth decreases, every \SystemNameO\ variant consistently surpasses reference models such as UDA and \SystemNameA, thereby establishing a stringent lower bound for lightweight consistency-regularized frameworks.
\SystemNameB intensifies regularisation through adversarial training and thereby achieves superior performance relative to \SystemNameO. Concurrently, \SystemNameO compresses the fine-tuning parameter count by a factor of 909 when a 12-layer BERT teacher distils into a 2-layer student, and the resultant inference time increases by only 0.46 second relative to \SystemNameB.

\begin{figure*}[tb]
\centering
\hspace{0.2mm}
\begin{minipage}[b]{.49\textwidth}
  \centering
  \includegraphics[width=1.0\linewidth]{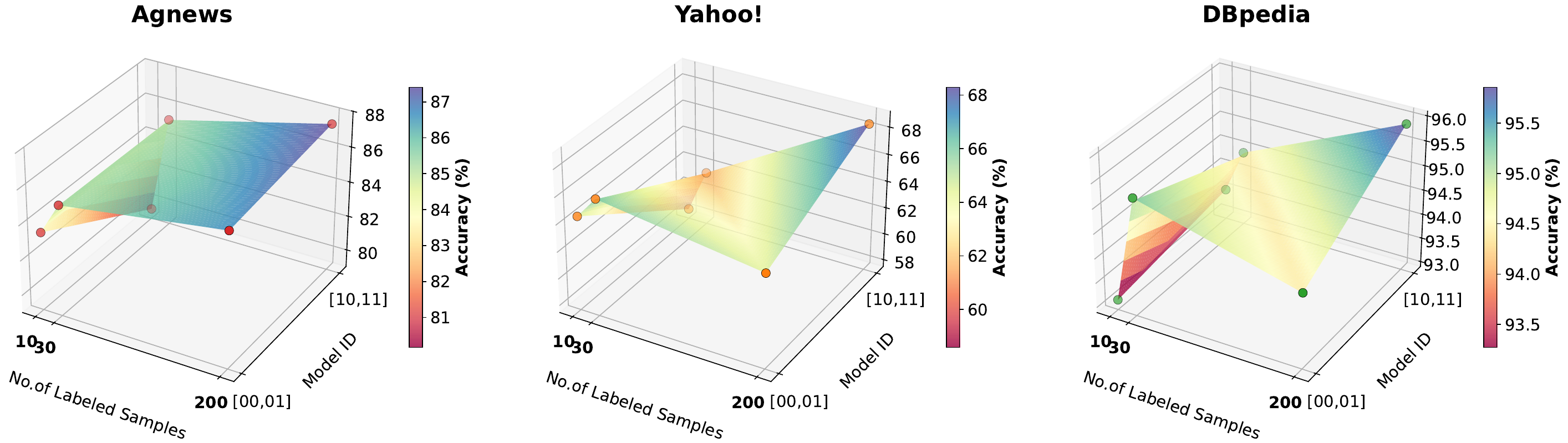}
  \\
  (a) {\scriptsize \SystemNameO with 2 Students Cohort}
\end{minipage}
\begin{minipage}[b]{.49\textwidth}
  \centering
  \includegraphics[width=1.0\linewidth]{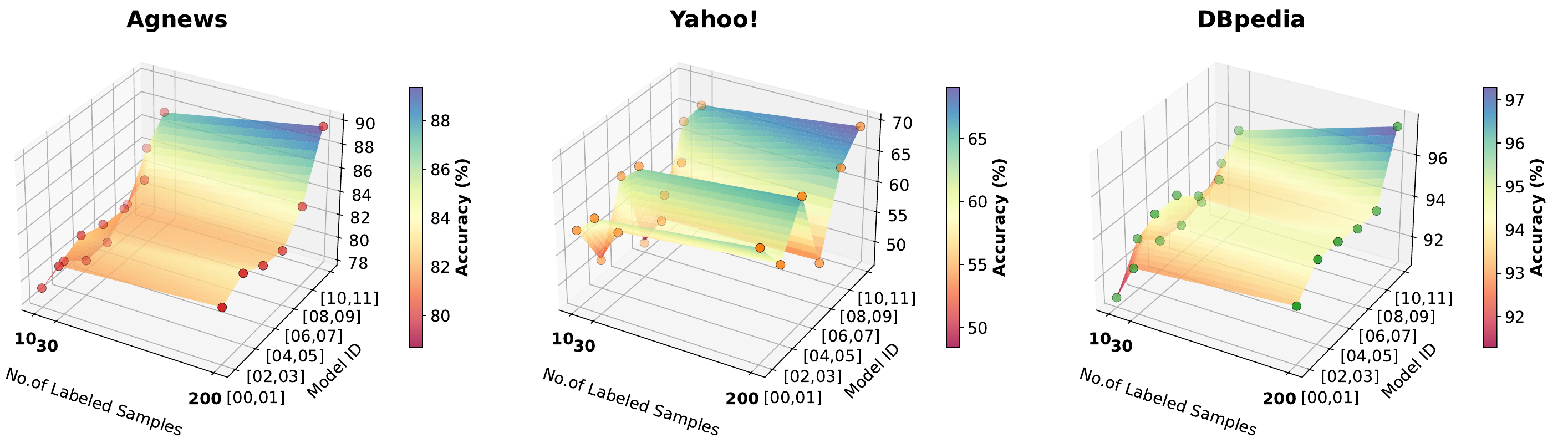}
  \\
  (b) {\scriptsize \SystemNameO with 6 Students Cohort}
\end{minipage}
\hspace{-0.1mm}
\caption{Accuracy Surface of Student Peers Across labeled-Sample Counts. X-axis: labeled-sample count. Y-axis: model ID. Z-axis: accuracy. Scatter points show student peers performance under equal labelling budgets with surface obtained by linear interpolation.}
\label{framework_comparative}
\end{figure*}

Table~\ref{tab:performance_comparison} consolidates eight-task metrics and positions \SystemNameO among eighteen baselines.
\SystemNameO (S\(^{\rm MBERT-A4}\)) achieves the highest mean accuracy (63.46\%) within the lightweight cohort, lagging the 109.48 M-parameter SoftMatch-BERT baseline (67.69\%) by only 4.23 percentage points while employing <0.5\% of its trainable parameters.
The shallower variants SMBERT-A2 and SMBERT-B2 remain stable (58.15\% and 55.75\%), indicating that depth reduction is gracefully absorbed by the cohort-wise consistency objective.  
\SystemNameO achieves competitive results across all corpora, underscoring the efficacy of parameter-efficient peer distillation under extreme label scarcity. Overall, the table demonstrates that \SystemNameO delivers a favourable accuracy-parameter trade-off: compressing the student to four layers halves inference parameters (101 M) yet preserves competitive performance, while further reduction to two layers still surpasses most lightweight SSL rivals.

\subsection{Qualitative Analysis}
This section qualitatively evaluates student-cohort expansion and contrasts two lightweight fine-tuning strategies: Embedding Frozen and Only-bias Tuning.  
Embedding Frozen is adopted because the embedding layer accounts for the predominant share of the fine-tunable parameter budget.
Only-bias Tuning is chosen because it adjusts only the bias terms and is empirically validated as the minimal-parameter instantiation of parameter-efficient training.

\subsubsection{Effect of More Student Peers}
We now examine how \SystemNameO scales as the cohort expands.
Each student receives individual supervision from every peer, so the pairwise teaching density grows with cohort size. Figure~\ref{framework_comparative} shows that enlarging the cohort to four students uniformly elevates the accuracy of every member, evidencing heightened generalisation under expanding peer perturbations.
These gains confirm that complementary student views constitute to attain stable performance despite model compression, scarce labels and few-parameter training.

\subsubsection{Effect of More Student Layers}

Table~\ref{main_text_classification} shows that, for the MBERT backbone, shrinking the student from 13 to 4 layers degrades the mean accuracy only marginally, from 82.50 to 81.57.  
Analogously, for the BERT backbone, reducing the student from 6 to 4 and finally to 2 layers yields 80.85, 79.67 and 79.26, respectively; hence depth compression preserves stability and avoids abrupt performance drops.  
Concretely, the MBERT student contracts from 212.14M to 101.71M inference parameters—an approximate halving—while the trainable count diminishes by merely 0.23M.
With only 10 labeled instances, the minimal MBERT and minimal BERT configurations outperform their larger counterparts; heavier parameter freezing and the reduced parameter budget jointly suppress variance and stabilise optimisation under extreme label scarcity.

\subsubsection{Effect of Embedding Frozen}


During our experiments, we found that when the number of layers in the student network of \SystemNameO is very small, reasonable settings are required for training, especially with less labeled data, as it is prone to overfitting. Therefore, we chose to freeze the embedding layer. Undoubtedly, freezing these parameters can reduce training overhead. For example, in the embedding layer, frozen parameters do not require gradient computation, which reduces the amount of backpropagation computation and significantly improves training efficiency. 
We conducted extensive visual analysis to confirm that freezing embeddings is indeed beneficial to model performance and effectiveness, as shown in Figure~\ref{framework_results}. For instance, in the case of only 10 labeled data points in each of the three datasets depicted in the three sub-figures, freezing often yields better results than not freezing. 
In fact, there is a general consensus~\cite{abs-2304-04662}\cite{ZhangRHLZ24},\cite{abs-2505-15696} that the embedding layer of Transformer typically retains general features and is suitable for being frozen.
When there is limited training data, freeze the parameters of the pre-trained model to preserve more of the initial general representation ability of the teacher model in \SystemNameO. After freezing, its feature extraction ability can be reused to avoid overfitting.


\section{Conclusion and Future Work}
We introduce \SystemNameO, a unified framework that simultaneously reduces model size through lightweight architectures, alleviates label scarcity via semi-supervised learning, and enables parameter-efficient training and inference with minimal labeled data. To the best of our knowledge, this is the first work to integrate these three objectives in a single system.
In future research, we will extend \SystemNameO to broader NLP benchmarks, including language understanding, machine reading comprehension, and text generation tasks. Additionally, we plan to adapt our approach to generative language models such as GPT and LLaMA, while continuing to explore its effectiveness with encoder-only models that remain prevalent in discriminative tasks such as classification.

\ifCLASSOPTIONcompsoc
  \section*{Acknowledgments}
\else
  \section*{Acknowledgment}
\fi




\ifCLASSOPTIONcaptionsoff
  \newpage
\fi




\normalem

{\footnotesize 
\bibliographystyle{ieeetr}
\bibliography{IEEEabrv}

}
\end{document}